\newif\ifconfver
\confvertrue        
\ifconfver
\documentclass[10pt,twocolumn,twoside]{IEEEtran}
\else
\documentclass[11pt,draftcls,onecolumn]{IEEEtran}
\fi
\usepackage{makecell}
\usepackage{bbding}
\usepackage{calc,amsfonts,amssymb,amsmath,bm,url,color,theorem,graphicx,cite}
\usepackage{psfrag,subfigure,float}
\usepackage{multirow,subfigure,amsmath,epsfig,amsfonts,amssymb,psfig,graphics,psfrag,theorem,calc,subfigure,url,bm,cite}
\usepackage{stfloats,hyperref}  
\usepackage{algorithm,algorithmic}
\usepackage{diagbox} 
\usepackage{hhline}
\usepackage{caption}
\usepackage{comment}
\usepackage{fancyhdr}
\usepackage{mathtools}

\makeatletter
\def\multilimits@{\bgroup
	\Let@
	\restore@math@cr
	\default@tag
	\baselineskip\fontdimen10 \scriptfont\tw@
	\advance\baselineskip\fontdimen12 \scriptfont\tw@
	\lineskip\thr@@\fontdimen8 \scriptfont\thr@@
	\lineskiplimit\lineskip
	\vbox\bgroup\ialign\bgroup\hfil$\m@th\scriptstyle{##}$\hfil\crcr}
\def\Sb{_\multilimits@}
\def\endSb{\crcr\egroup\egroup\egroup}
\makeatother
{\begin{list}{}%
		{\setlength{\rightmargin}{0pc}%
			\setlength{\leftmargin}{1pc} }} %
	{\end{list}}
\newlength{\twidth}
\ifconfver
\else
\fi
\definecolor{orange}{RGB}{255,107,0}

\theorembodyfont{\rmfamily}

{\begin{list}{}{
			\settowidth{\labelwidth}{\mbox{\textnormal{#1}}}%
			\setlength{\leftmargin}{\labelwidth+\labelsep}}}%
	{\end{list}}
\newcommand\bA{\ensuremath{{\bm A}}}
\newcommand\bB{\ensuremath{{\bm B}}}

\newcommand\bG{\ensuremath{{\bm G}}}
\newcommand\bH{\ensuremath{{\bm H}}}

\newcommand\bK{\ensuremath{{\bm K}}}

\newcommand\bQ{\ensuremath{{\bm Q}}}

\newcommand\bS{\ensuremath{{\bm S}}}
\newcommand\bT{\ensuremath{{\bm T}}}

\newcommand\bV{\ensuremath{{\bm V}}}

\newcommand\bX{\ensuremath{{\bm X}}}
\newcommand\bY{\ensuremath{{\bm Y}}}

\newcommand\bb{\ensuremath{{\bm b}}}

\newcommand\bd{\ensuremath{{\bm d}}}

\newcommand\bt{\ensuremath{{\bm t}}}

\newcommand\bx{\ensuremath{{\bm x}}}
\newcommand\by{\ensuremath{{\bm y}}}

\definecolor{orange}{RGB}{255,107,0}

\ifconfver
\author{Po-Wei Tang,~\IEEEmembership{Student Member,~IEEE}, Chia-Hsiang Lin,~\IEEEmembership{Member,~IEEE}, and Yangrui Liu,~\IEEEmembership{Student~Member,~IEEE}}
\title{Transformer-Driven Inverse Problem Transform For Fast Blind Hyperspectral Image Dehazing

\thanks{This study was supported partly by the EINSTEIN Program of National Science and Technology Council (NSTC), Taiwan, under Grant MOST 111-2636-E-006-028; 
and partly by the Emerging Young Scholar Program of NSTC, Taiwan, under Grant NSTC 112-2628-E-006-017.
We thank the Center for Data Science at NCKU, and National Center for High-performance Computing (NCHC) for providing the computing resources.}
\thanks{\textit{(Corresponding author: Chia-Hsiang Lin)}}
    \thanks{P.-W. Tang is with the Institute of Computer and Communication Engineering, Department of Electrical Engineering, National Cheng Kung University, Tainan, Taiwan (R.O.C.) 
		(e-mail:  q38091526@gs.ncku.edu.tw).}
    \thanks{C.-H. Lin is with the Department of Electrical Engineering, and with the Miin Wu School of Computing, National Cheng Kung University, Tainan, Taiwan (R.O.C.) 
(e-mail: chiahsiang.steven.lin@gmail.com).}
\thanks{Y. Liu is with the Institute of Computer and Communication Engineering, Department of Electrical Engineering, National Cheng Kung University, Tainan, Taiwan (R.O.C.) 
		(e-mail:  q38103501@gs.ncku.edu.tw).}
}%

\fancypagestyle{IEEEtitlepagestyle}{
	\fancyhf{} 
	\fancyhead[LE,RO]{\thepage} 
	\fancyfoot[C]{\scriptsize © 2024 IEEE. Personal use of this material is permitted. Permission from IEEE must be obtained for all other uses, in any current or future media, including reprinting/republishing this material for advertising or promotional purposes, creating new collective works, for resale or redistribution to servers or lists, or reuse of any copyrighted component of this work in other works. This is the accepted version of the manuscript. The final, published version is available at IEEE Xplore via https://doi.org/10.1109/TGRS.2024.3349479.}
	
}
\else
\fi
\begin{document}
\bibliographystyle{IEEEtran}
\maketitle
\ifconfver 
\else 
\vspace{-0.5cm}
\fi
\begin{abstract}
Hyperspectral dehazing (HyDHZ) has become a crucial signal processing technology to facilitate the subsequent identification and classification tasks, as the AVIRIS data portal reports a massive portion of haze-corrupted areas in typical hyperspectral remote sensing images.
{The idea of inverse problem transform (IPT) has been proposed in recent remote sensing literature in order to reformulate a hardly tractable inverse problem (e.g., HyDHZ) into a relatively simple one.}
%
%
Considering the emerging spectral super-resolution (SSR) technique, which spectrally upsamples multispectral data to hyperspectral data, we aim to solve the challenging HyDHZ problem by reformulating it as an SSR problem.
Roughly speaking, the proposed algorithm first automatically selects some uncorrupted/informative spectral bands, from which SSR is applied to spectrally upsample the selected bands in the feature space, thereby obtaining a clean hyperspectral image (HSI).
The clean HSI is then further refined by a deep transformer network to obtain the final dehazed HSI, where a global attention mechanism is designed to capture non-local information.
There are very few HyDHZ works in existing literature, and this paper introduces the powerful spatial-spectral transformer into HyDHZ for the first time.
Remarkably, the proposed transformer-driven IPT-based HyDHZ (T$^2$HyDHZ) is a blind algorithm without requiring the user to manually select the corrupted region.
Extensive experiments demonstrate the superiority of T$^2$HyDHZ with less color distortion.
%
%
%
%
%
%
\\

\end{abstract}

\ifconfver \else \vspace{-0.0cm}\fi
\ifconfver \else \vspace{-0.5cm}\fi
%
\ifconfver \else  \fi
\section{Introduction}\label{sec:intro}
\subsection{Overview}\label{sec:overview}
%
{Hyperspectral image (HSI) data contains a wealth of spectral information from visible to short-wave infrared (VSWIR) regions with wavelengths between 400 and 2500 nm, which has led to impactful development and influence \cite{BPCSNC2013,reviewaccess14,lin2021all}.}
%
{Specifically, HSI has been endowed with extensive applications in remote sensing, including land cover classification,  change detection, environment monitoring, and mangrove mapping \cite{Hongtgrs2021,Houtgrs2022,codehcd,underwood2006mapping,codemm}.}
%
%
{However, these practical applications \cite{CVXbookCLL2016} may face a significant challenge --- the scarcity of applicable data due to atmospheric effects on image quality \cite{narasimhan2002vision}.}
As stated in the Airborne Visible/Infrared Imaging Spectrometer (AVIRIS) data portal from 2006 to 2021, only a small amount of the acquired HSIs are completely free from atmospheric interference, while the remaining are affected by atmospheric particle absorption and the reflection of airlight \cite{GuoTGRS2021, XudongTGRS2021}, such as haze or cloud coverage.
%
%
{To address the lack of data, one potential solution is to re-photograph the area during better atmospheric conditions. 
However, this approach could result in additional expenses and time costs. Moreover, it might not be feasible if the site is frequently impacted by haze pollution.}

{As more practical solutions, various} dehazing algorithms have been developed to restore hazy images, which can be broadly classified into three categories: dark-object subtraction (DOS) model-based \cite{DOS2022}, atmospheric scattering (AS) model-based \cite{asmodel2000}, and neural network (NN) model-based methods \cite{Caidehazenettip2016}.
%
%
Before comprehensively introducing these approaches below, {it is essential to} remark that the hyperspectral dehazing (HyDHZ) problem is similar {to} but different from the hyperspectral inpainting problem {\cite{LinTNNLS2023,hyperqueen}}.
%
%
First, the DOS model assumes that the clean image $\bX$ can be retrieved by directly eliminating the haze component $\bH$ from the original hazy image $\bY$, defined as follows:
\begin{equation} \label{eq:dos}
\bY= \bX + \bH,
\end{equation}
where $\bY$, $\bX$, and $\bH$ $\in \mathbb{R}^{H\times W \times C}$ with the size of height $H$, width $W$, and bands $C$, respectively;
$\mathbb{R}^{H\times W \times C}$ is the $H\times W \times C$-dimensional real-valued 3-way tensor space.
Second, the AS model mathematically describes the absorption and the scattering effect of light through atmospheric particles or water droplets, as defined in recent literature \cite{hetpami2010,GuoTGRS2021}:
\begin{equation}
\label{eq:AS}
 \bY=\bX \odot \bT+\bA \odot (\textbf1_{H\times W \times C}-\bT),
 \end{equation}
where $\bT$ and $\bA$  $\in \mathbb{R}^{H\times W \times C}$ represent the medium transmission and the global atmospheric light \cite[Figure 2]{TanCVPR2008}, respectively;
$\odot$ denotes the Hadamard product;
$\textbf1_{H\times W \times C}$ denotes the tensor with all-one elements in the given dimension.
In summary, \eqref{eq:AS} depicts that a hazy image comprises the direct attenuation of the radiance scene (i.e., term $\bX \odot \bT$) and the ambient light present in the atmosphere (i.e., term $\bA \odot (\textbf1_{H\times W \times C}-\bT)$) according to the Beer-Lambert law \cite{TanCVPR2008}.
Third, the NN model is designed to generate an efficient non-linear function $f_{\theta}(\cdot)$ by iteratively updating the network parameters $\theta$ with the back-propagation algorithm \cite{bp}, thereby learning the mapping from the corrupted HSI to its clean counterpart.
%
%
Compared to the DOS model, {the AS model further considers the AS effects rather than just treating the haze component as a noise component to be removed.} 
{Thus, since} the AS model considers the natural physical meaning, it has been utilized to simulate the testing hazy data in recent literature \cite{guo2020rsdehazenet,ma2022spectral}.
However, these models are limited to representing only linear relationships.
%
%
{By contrast, the NN model offers another perspective to solve the dehazing problem by capturing non-linear relationships of atmospheric effects through well-developed deep-learning technology.}

\subsection{Related Work}\label{sec:relatedwork}
%
%
%
{Due to the limited research on hyperspectral dehazing (HyDHZ), this section will further explore existing multispectral dehazing (MuDHZ) methods with recent dehazing algorithms, as categorized in sequence.} 
%
%

{Inspired by the DOS model in \eqref{eq:dos}, Kang \textit{et al.}\cite{XudongTGRS2021} proposed a two-step HyDHZ algorithm called fog model-based HSI defogging (FHD).
First, FHD calculates a fog intensity map by subtracting the average of visible bands from the average of infrared bands and then estimates corresponding haze abundance by manually selecting the same land cover with different haze levels (i.e., hazy and clean land covers). 
Second, FHD eliminates the haze component from the hazy HSI by utilizing the fog intensity and associated haze abundance maps \cite{lin2016fast}.
%
%
Apart from the DOS-based method, Guo \textit{et al.} \cite{GuoTGRS2021} proposed an optimized-based AS model (MDOAS) for MuDHZ, which considers the relationship between the scattering coefficient and wavelength to estimate the transmission rate and atmosphere light for obtaining a recovered dehazed result.
With the development of NN, Qin \textit{et al.} \cite{QinJSTARS2018} proposed a parallel-wise cascaded convolutional neural network (CNN) with a residual structure to extract and fuse multi-scale features for restoring hazy Landsat-8 operational land imager (OLI) multispectral images. 
Guo \textit{et al.} \cite{guo2020rsdehazenet} proposed another residual learning-based network, RSDehazeNet, further considers both local and global features by utilizing the channel attention mechanism to focus on the critical informative bands, thereby improving the accuracy and robustness of the image recovery process.
Furthermore, Ma \textit{et al.} \cite{ma2022spectral} proposed a classical residual network with spectral grouping and multi-scale attention modules, SG-Net, to recover the hazy HSI successfully.
It is worth noting that Song \textit{et al.} \cite{song2022rethinking} proposed GUNet, a variant UNet combining residual blocks with gating mechanisms and selective kernels, which achieved satisfactory dehazing results on the RS-Haze dataset.
Besides, similar to the concept of FHD, Zi \textit{et al.} \cite{MTCRN2021jstar} proposed a UNet-based multispectral thin cloud removal network (MTCRN) to estimate a more accurate thin cloud thickness map as a reference and then calculate the thickness coefficients of each band with a mathematical imaging model, effectively removing thin clouds from Landsat-8 OLI multispectral images.
From another perspective, Gan \textit{et al.}\cite{gan2016dehazing} proposed a novel unmixing-based HyDHZ approach, termed as HUD.
Specifically, HUD treats haze in the HSI as one of the endmembers in the unmixing model {\cite{HISUN,lin2015identifiability}}. 
To separate haze information from the observed hazy image, HUD adopted a fast algorithm for linearly unmixing (FUN)  \cite{FUNtgrs15} for reconstructing the image with sum-to-one and non-negative assumptions.}
%
%
%
%
Besides, Tang \textit{et al.} proposed CODE-HD  \cite{Tang2022dehazing} based on a radically new CODE theory \cite{LinTGRS2021} by combining convex optimization (CO) and deep learning (DE) to recover hazy HSIs.
{CODE theory was initially developed to solve the hyperspectral inpainting problem in scenarios where only a small dataset is available.} 
Benefitting from the CODE framework, CODE-HD employs the $\bQ$-norm as a deep regularizer to extract spatial information from a rough DE solution, which is trained with limited small data.
By further considering additional {temporal} information, Ma \textit{et al.} \cite{MA2022113012} proposed a multi-temporal HyDHZ algorithm called TIIN, which fuses the {dehazed} result using haze-free information captured at different dates. 
Nevertheless, this approach incurs extra time for data acquisition.

\subsection{Motivation and Theory}\label{sec:motivation}

However, potential limitations exist in the benchmark MuDHZ and HyDHZ methods (i.e., MDOAS and FHD), which depend on laboriously handcrafted parameters {and settings} that are challenging to be {well-adjusted} for general users.
Therefore, these hindrances inspire us to ask the question:
\textit{Is it possible to design a fast{,} high-performance HyDHZ algorithm without requiring the users to adjust handcrafted parameters or to identify the haze-corrupted regions manually?}
{This is affirmative and has driven us to design a non-parameter end-to-end network for HyDHZ.
Inspired by the natural characteristic of transmission rate over wavelength, we design an interpretable network that ingeniously utilizes the physical meaning discussed in Section \ref{sec:ipt}.
%
%
Specifically, from the physical perspective, we transform a hardly tractable inverse problem (e.g., HyDHZ) into a relatively well-studied one as the inverse problem transform IPT-motivated method, which aligns with natural properties and physical meanings, detailed in Section \ref{sec:ipt}.}
%
%
\begin{figure}[t]
    \begin{center}
        \includegraphics[width=0.5\textwidth] {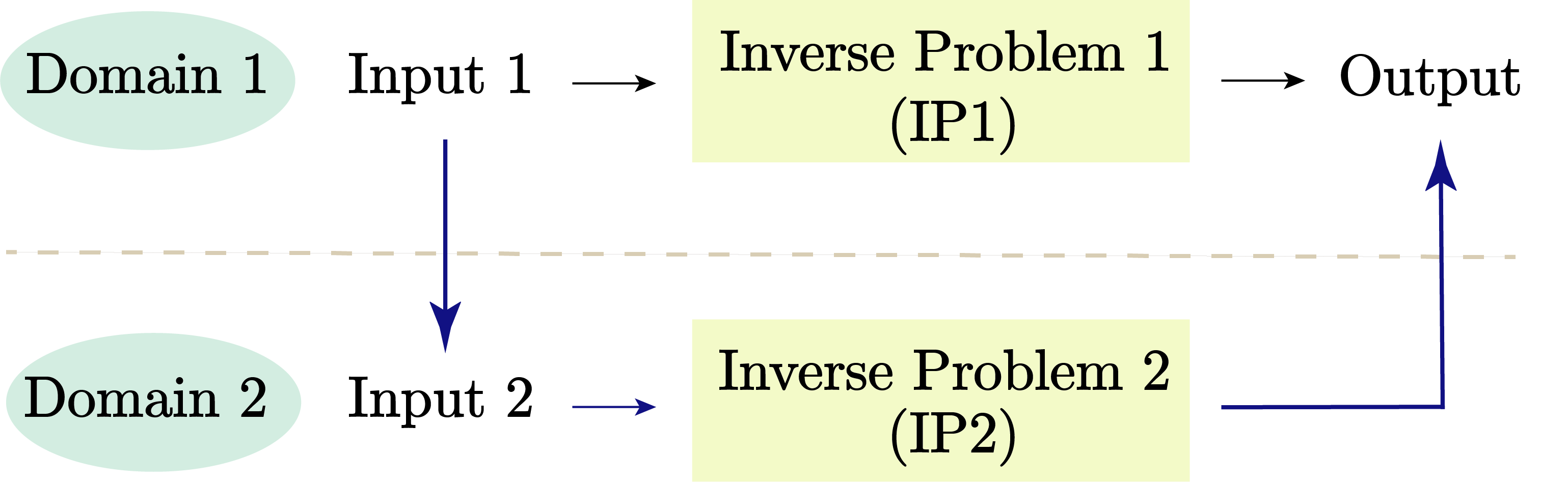}
    \end{center}
    \caption{Graphical illustration of the idea of inverse problem transform (IPT), which solves the more challenging IP1 in another domain wherein a relatively easier IP2 can be more efficiently addressed.}
    \label{fig:ipt}
\end{figure}
%
%
{The idea behind IPT \cite{iptwhsipers} is to simplify a complex inverse problem (IP1) into a more manageable inverse problem (IP2) by leveraging physical interpretations.}
As graphically illustrated in Figure \ref{fig:ipt}, the IPT theory is practically very helpful if the following three conditions hold true: 
\begin{enumerate}
\item {B}oth IP1 and IP2 have the same target output{.}

\vspace{0.1cm}

\item {T}he input of IP1 (i.e., input 1) can be transformed into the input of IP2 (i.e., input 2){.}

\vspace{0.1cm}

\item IP2 has been more extensively studied (or can be more easily solved) than IP1.
\end{enumerate}
%
%
%

In recent years, the spectral super-resolution (SSR) problem has drawn considerable {attention \cite{Linnature2023,Hetnnls2022,Hangtip2021,Yitgrs2019}}, while there are very few works on HyDHZ (cf. condition 3).
{Most importantly, utilizing the abovementioned natural physical meaning leads to an interpretable and feasible model design.}
So, we aim to solve the highly challenging HyDHZ problem (i.e., IP1) by transforming it into the SSR problem (i.e., IP2).
Note that condition 1 and condition 2 are also satisfied, as explained below.
{In} condition 1, SSR aims to construct a complete HSI (from an input multispectral image (MSI)), whereas HyDHZ {seeks} to restore a complete HSI (from an input haze-corrupted HSI); in other words, both SSR and HyDHZ have the same output.
{Regarding} condition 2, the input of IP1 (i.e., haze-corrupted HSI) can be transformed into the input of IP2 (i.e., clean/informative MSI) through the proposed band selection technique{.} 
{S}pecifically, our method will automatically select those clean/informative bands from the original haze-corrupted HSI {to form a clean/informative MSI, which is then spectrally super-resolved} by SSR to obtain the target clean/informative HSI.
{As all three conditions are fulfilled, we implement an IPT-motivated deep end-to-end transformer network.}

\begin{figure*}[t]
    \begin{center}
        \includegraphics[width=1\textwidth] {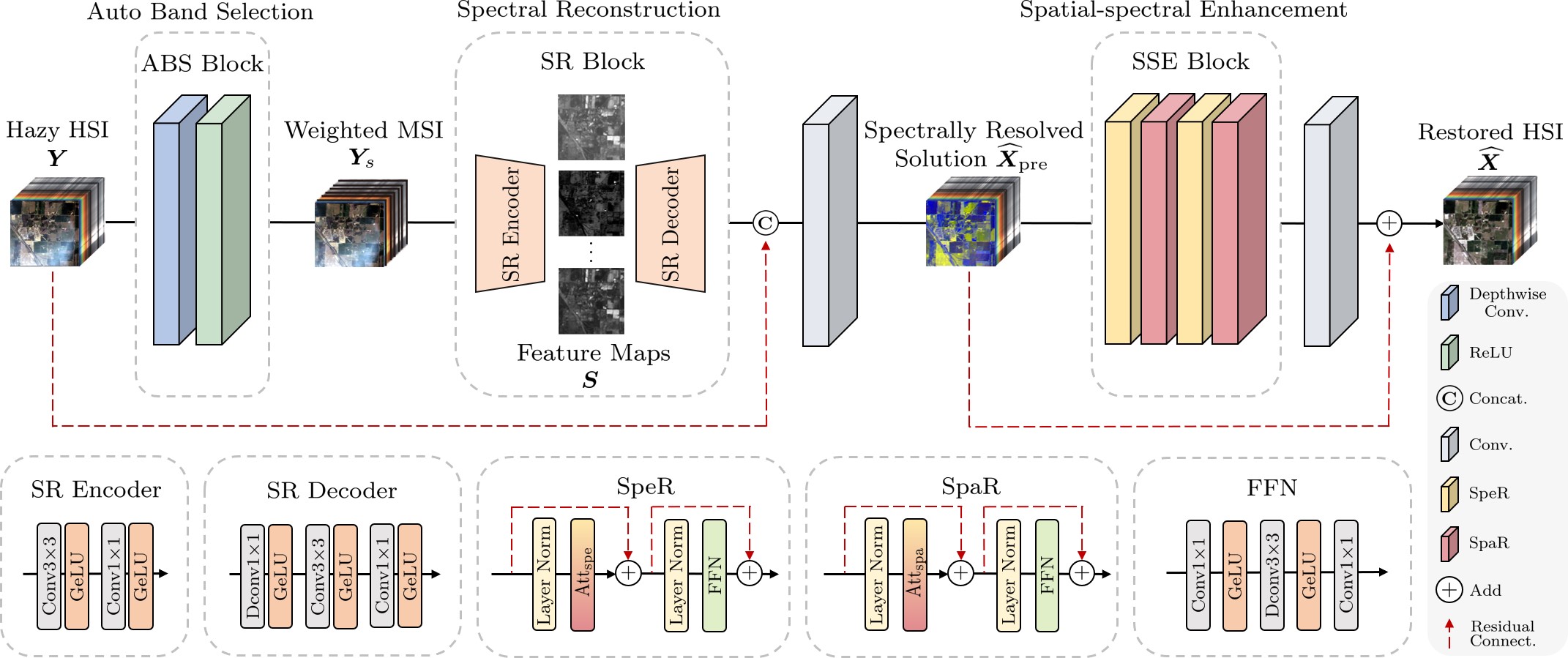}
    \end{center}
    \caption{Graphical illustration of the proposed T$^2$HyDHZ network structure, which consists of three fundamental blocks, namely the auto band selection (ABS), spectral reconstruction (SR), and spectral-spatial enhancement (SSE) blocks. 
Particularly, the SSE block comprises spectral refinement (SpeR) and spatial refinement (SpaR), which adopt global self-attention in specific spectral and spatial dimensions (i.e., $\text{Att}_{\text{spe}}$ and $\text{Att}_{\text{spa}}$), respectively.
Additionally, a feed-forward network (FFN) is employed to obtain higher-level features after the global attention mechanism in the refinement step.}
    \label{fig:framework}
\end{figure*}
\subsection{Contributions}\label{sec:contribution}

The proposed transformer-driven IPT-based HyDHZ (T$^2$HyDHZ) is a blind algorithm without requiring the user to manually select
the corrupted region{; while it is very fast as both transform and SSR stages can be efficiently implemented.}
T$^2$HyDHZ {makes} notable advances in HyDHZ, as evidenced by its outstanding dehazing results in both qualitative and quantitative {analyses}. 
The major contributions of this article are summarized as follows:

\begin{itemize}
\item A user-friendly and fully automatic NN model is proposed for HyDHZ, obviating the necessity for manual parameter tuning.
Furthermore, T$^2$HyDHZ is a blind algorithm, meaning that the users do not need to manually label those haze-corrupted regions, which is a daunting task required by other benchmark methods.

\vspace{0.1cm}

\item By adopting the IPT concept, the HyDHZ problem is innovatively reformulated as {an} SSR problem and then effectively solved. 
The T$^2$HyDHZ network contains i) an auto band selection (ABS) module that selects informative bands from hazy hyperspectral bands to form a clean MSI, ii) a spectral reconstruction (SR) module that generates a spectrally super-resolved solution, and iii) a refinement module that improves spectral-spatial details in the final outcome.
The three fundamental yet effective modules yield state-of-the-art HyDHZ performance both quantitatively and qualitatively.

\vspace{0.1cm}

\item For the first time, the spatial-spectral transformer module is introduced into the HyDHZ problem, which allows us to capture global relationships using a global attention mechanism.
{Besides,} the overall computational time is much faster than existing benchmark methods.
\end{itemize}

In the remainder of this article, we {provide} a comprehensive overview of our proposed method for tackling the challenging high-dimensional HyDHZ problem in remote sensing.
In Section \ref{sec:theory}, we will present {an in-depth explanation} of the proposed T$^2$HyDHZ algorithm.
In Section \ref{sec:experiment}, extensive experiments with synthetic and natural hazy data are presented and analyzed; particularly, ablation {studies are} conducted to demonstrate the effectiveness of {each} module of the proposed framework, {to assess the efficacy of the $L_\text{sparsity}$ loss, and to evaluate the benefits of the concatenation component within the framework} in Section \ref{sec:ablation}.
Finally, we conclude our findings and insights in Section \ref{sec:conclusion}.
%
%
\section{Blind Hyperspectral Image Dehazing Via Transformer-Driven Inverse Problem Transform}\label{sec:theory}

In this section,  we propose a novel framework to solve the HyDHZ problem {motivated by} the IPT theory \cite{iptwhsipers}.  
This framework offers an effective solution by transforming the challenging dehazing problem into an SSR problem. 
By this problem transformation, the issue of HyDHZ can be solved by performing SSR on informative bands of the hazy image.
As shown in Figure \ref{fig:framework}, the proposed end-to-end network consists of three modules: the auto band selection (ABS) block, the spectral reconstruction (SR) block, and the spectral-spatial enhancement (SSE) block.
In addition, the designed {network is a blind HyDHZ method, which eliminates the need for handcrafted parameter settings (i.e., manual parameter adjustment and haze/non-haze region selection)}, which are required in existing HyDHZ methods.
%
%

%

%
%
\subsection{Inverse Problem Transform (IPT) on HyDHZ Problem} \label{sec:ipt}
\begin{figure}[t]
	\begin{minipage}{0.242\linewidth}
		\centerline{\includegraphics[width=1\textwidth]{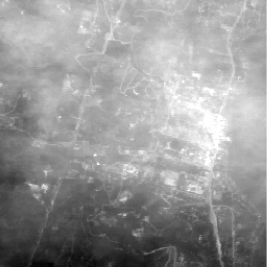}}
		\vspace{-0.1cm}
		\caption*{(a) 463nm}
	\end{minipage}
	\begin{minipage}{0.242\linewidth}
		\centerline{\includegraphics[width=1\textwidth]{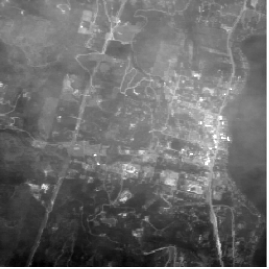}}
		\vspace{-0.1cm}
		\caption*{(b) 550nm}
	\end{minipage}
	\begin{minipage}{0.242\linewidth}
		\centerline{\includegraphics[width=1\textwidth]{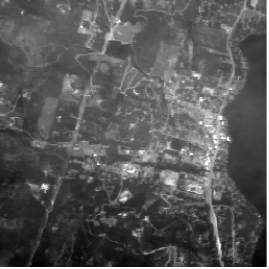}}
		\vspace{-0.1cm}
		\caption*{(c) 648nm}
	\end{minipage}
	\begin{minipage}{0.242\linewidth}
		\centerline{\includegraphics[width=1\textwidth]{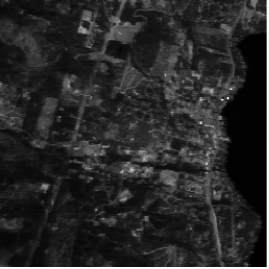}}
		\vspace{-0.1cm}
		\caption*{(d) 2027nm}
	\end{minipage}
        \begin{minipage}{0.242\linewidth}
		\vspace{0.1cm}\centerline{\includegraphics[width=1\textwidth]{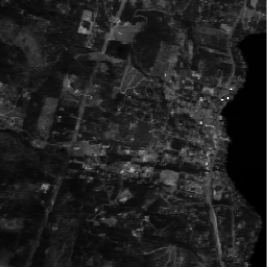}}
		\vspace{-0.1cm}
		\caption*{(e) 2117nm}
	\end{minipage}
        \begin{minipage}{0.242\linewidth}
		\vspace{0.1cm}\centerline{\includegraphics[width=1\textwidth]{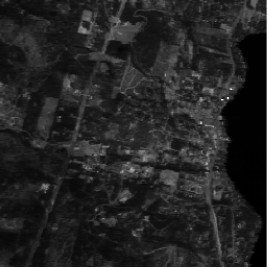}}
		\vspace{-0.1cm}
		\caption*{(f) 2177nm}
	\end{minipage}
        \begin{minipage}{0.242\linewidth}
		\vspace{0.1cm}\centerline{\includegraphics[width=1\textwidth]{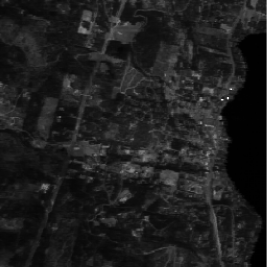}}
		\vspace{-0.1cm}
		\caption*{(g) 2277nm}
	\end{minipage}
	\begin{minipage}{0.242\linewidth}
		\vspace{0.1cm}\centerline{\includegraphics[width=1\textwidth]{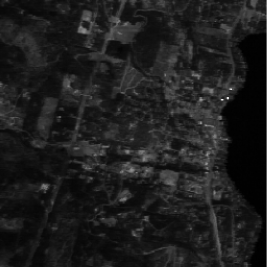}}
		\vspace{-0.1cm}
		\caption*{(h) 2366nm}
	\end{minipage}
 
	\caption{Gray-scale images show the pollution extent of different center wavelengths in the AVIRIS HSI data over Arizona, USA. (a)-(c) is in visible bands. (d)-(h) is in infrared bands. 
 }\label{fig:visible/infrared}
\end{figure}

{The concept of IPT stems from the physical characteristics and properties of authentic hazy images.
In recent literature \cite{GuoTGRS2021,XudongTGRS2021}, it has been observed} that the transmission rate is higher for longer wavelengths, implying that the impact of haze is smaller on spectral bands corresponding to longer wavelengths.
{To facilitate the comprehension of the readers, Figure \ref{fig:visible/infrared} displays the extent of pollution in various bands from a real hazy AVIRIS HSI over Arizona, USA.}
One can see that the pollution extent decreases as the wavelength increases.
%
{Accordingly, we aim to design an interpretable model that utilizes those informative bands with less pollution.
Motivated by the idea behind IPT, if we can identify high-quality bands in a hazy image, the HyDHZ problem (i.e., IP1) can be reformulated as the SSR problem (i.e., IP2) with exact physical significance, implying both IPs have the same objective to recover target output (cf. condition 1 in Section \ref{sec:intro}).
As demonstrated in Figure \ref{fig:ipt}, we provide an illustration to enhance comprehension of the IPT concept.
Please refer to Section \ref{sec:intro} to get a better sense of applying the IPT theory with three basic conditions.
%
%
%
IPT was initially designed to solve a challenging IP1 in a more friendly problem domain wherein another IP2 can be resolved more efficiently \cite{iptwhsipers}.
By drawing inspiration from the abovementioned real physical meaning and motivated by IPT theory,  we present a novel method T$^2$HyDHZ to tackle the challenging HyDHZ problem (i.e., IP1) by solving the more well-studied spectral super-resolution (SSR) problem (i.e., IP2).}

%

%

As the SSR problem has drawn considerable attention {\cite{Linnature2023,Hetnnls2022,Hangtip2021,Yitgrs2019}} (cf. condition 3 in Section \ref{sec:intro}), we aim to solve the highly challenging HyDHZ problem (i.e., IP1) by transforming it into the SSR problem (i.e., IP2).
{Most importantly, utilizing the abovementioned natural physical meaning leads to an interpretable and feasible model design.}
Note that condition 1 and condition 2 are also satisfied.
For condition 1, SSR aims to construct a complete HSI (from an input MSI), whereas HyDHZ {seeks} to restore a complete HSI (from an input haze-corrupted HSI); in other words, both SSR and HyDHZ have the same output (cf. condition 1 in Section \ref{sec:intro}).
Moreover, the input of IP1 (i.e., haze-corrupted HSI) can be transformed into the input of IP2 (i.e., clean/informative MSI) through the proposed band selection technique. 
{S}pecifically, our method will automatically {select} those clean/informative bands from the original haze-corrupted HSI, allowing us to form a clean/informative MSI that will then be spectrally super-resolved to obtain the target clean/informative HSI (cf. condition 2 in Section \ref{sec:intro}).

{Unlike the original IPT framework, we design a modified version of IPT that maximizes available information by further adopting a shortcut connection with hazy HSI $\bY$ to achieve optimal performance, which is evaluated in Section \ref{sec:ablation}.
In summary, we propose a physical-meaning-driven and IPT-motivated approach utilizing a deep end-to-end transformer network.}
By reframing the HyDHZ problem, {we can offer an efficient solution with a different perspective using an interpretable and customized neural network model based on SSR.}
The proposed deep transformer {only needs to learn} the non-linear relationship between spectral bands, which is comparatively less complex than solving IP1 and leads to {a} much faster computational speed thanks to the IPT theory.

\subsection{Auto Band Selection (ABS)} \label{sec:abs}

In Section \ref{sec:abs}, we will introduce how to obtain the input 2 (i.e., {clean/informative MSI}) for solving IP2 (i.e., SSR problem).
The ABS block is designed to automatically select {clean/information MSI}, providing enough spatial information for the subsequent SR step. 
As mentioned in Section \ref{sec:ipt}, light with longer wavelengths exhibits higher transmission rates. 
{This natural fact} implies that the ground information of these bands can more easily penetrate through the haze and hence be captured by sensors, thereby generating clearer images.  
Building upon this physical property, a feasible strategy is to directly select the bands of longer wavelengths as the input for subsequent SSR.
However, this strategy may result in overlooking other bands that could potentially offer valuable information for SSR.
With this concern, we opt for an alternative design approach: \textit{integrating band selection within the overall network architecture to create an end-to-end model.} 
This design brings {the} advantage of providing the model greater flexibility for automatic adjustment based on the datasets, leading to an overall improved model fitting \cite{NIPS2015_8fb21ee7}.

In the ABS block, we devise a novel mechanism that combines channel attention and additional filtering, named as the band attention mechanism, to select {clean/}informative bands and assign importance weights to each selected band.
Specifically, the proposed method first utilizes the channel attention mechanism to learn individual weights for each band of input data. 
Then, the weights are multiplied on each of the spectral bands, resulting in a weighted hyperspectral cube, thereby enabling the model to focus on more crucial and significant information. 
Finally, a rectified linear unit (ReLU) operator $\text{ReLU}(\cdot)$ \cite{nair2010rectified} is applied to filter the data weighted by the channel attention mechanism. 
%
%
{Thus, only positive-weighted bands are retained with ReLU filtering, while others become zero matrices.}
The channel attention mechanism is implemented by a $1 \times 1$ depthwise convolutional layer $f_{\theta_\text{D}}(\cdot)$, where $\theta_\text{D}$ defines the learning weights in $f_{\theta_\text{D}}(\cdot)$.  
To conclude, the procedure of the ABS block for selecting {clean/}informative bands is defined as follows:
\begin{align}
\bY_s=\text{ReLU}(f_{\theta_\text{D}}(\bY)),
\end{align}
where $\bY_s\in \mathbb{R}^{H\times W \times C}$ represents the clean/informative MSI consisting of weighted selected and zero-valued bands for end-to-end training.  
Through the above procedure, the selection of {clean/}informative bands can be effectively implemented.
{To briefly explain, the ABS process involves assigning weights to capture information and filtering out negative-valued bands, enabling the network to identify informative bands with positive weights.
This process will create a 3-D cube with the same size as the input HSI, including informative bands with positive values and uninformative bands with zero values.
If we neglect the zero-valued bands (i.e., uninformative bands), this 3-D cube can be viewed as a weighted MSI.}
\begin{figure}[t]
    \begin{center}
        \includegraphics[width=0.49\textwidth] {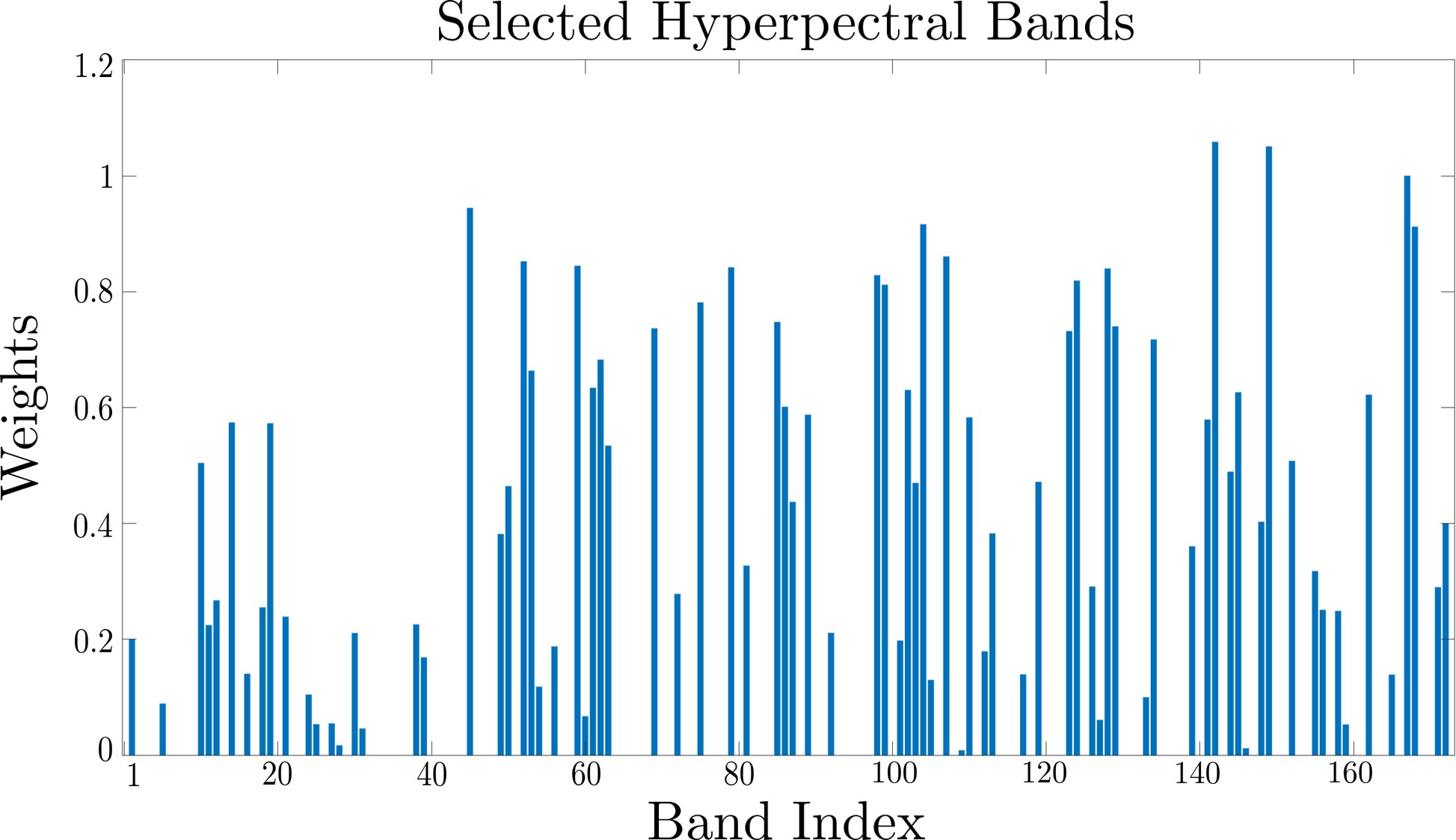}
    \end{center}
    \caption{Visual presentation of bands selected by the ABS block. 
    The bands are ordered according to their wavelengths, so the infrared bands are indexed by those $i>41$.
    One can see that those informative bands (i.e., with higher attention weights) do mostly concentrate on the relatively clean infrared region.}
    \label{fig:weight}
\end{figure}

{To gain a better understanding of the selected band by the ABS block, Figure \ref{fig:weight} illustrates the positive learning weights of $f_{\theta_\text{D}}(\cdot)$ $\in\mathbb{R}^{172}$ after being filtered with the ReLU operator.
Besides, the channel attention mechanism selects the most significant bands for the ABS module with positive weights, while unselected bands are set to zero.}
%
%
Notably, longer wavelength bands (i.e., the relatively clean infrared region) receive higher weights compared to shorter wavelength bands (i.e., the visible region). 
{Furthermore, since the neighboring bands of HSI have highly correlated information, the ABS block learns to select representative bands from adjacent regions, even in the long wavelength.
This result highly aligns with prior knowledge on transmission rate and band correlation, thereby demonstrating the physical meaning of the proposed ABS module in selecting clean/informative bands.}
%
%
In addition, the effectiveness of the proposed ABS module on the overall model will be double{-}verified using the ablation study in Section \ref{sec:ablation}.

\subsection{Spectral Reconstruction (SR)}\label{sec:sr}

After successfully acquiring {clean/}informative bands selected by the ABS block, the next step is to reconstruct a  spectrally {super-resolved} solution $\bX_{\text{pre}}$ by SSR.
%
%
Hence, for generating the {target clean/informative HSI} from the specified bands selected by the ABS block, we adopt an encoder-decoder structure and the Gaussian error linear units (GeLU) \cite{Hendrycks2016GaussianEL} to learn the representations in the embedding space, defined as follows:
\begin{align}
\text{Encoder}(\mathbf{X})&=\sigma(f_{\theta_2}(\sigma(f_{\theta_1}(\mathbf{X})))),
\\
\text{Decoder}(\mathbf{X})&=\sigma(f_{\theta_5}(\sigma(f_{\theta_4}(\sigma(f_{\theta_3}(\mathbf{X})))))),
\end{align}
where $\sigma(\cdot)$ denotes the GeLU activation function; $f_{\theta_i}(\cdot)$ is the $i$th convolutional layers of the SR block, as illustrated in Figure \ref{fig:framework}; $\theta_i$ means the corresponding learning weights in $f_{\theta_i}(\cdot)$; $\mathbf{X}$ represents the 3-D tensor as input.
\begin{figure*}[t]
    \begin{center}
   \includegraphics[width=0.99\textwidth] {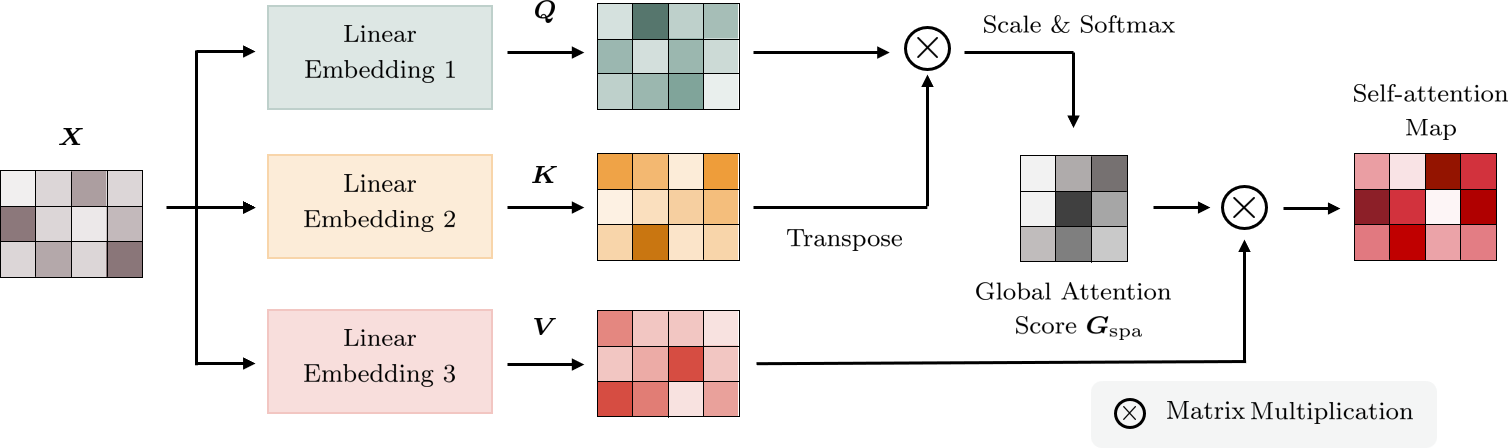}
    \end{center}
    \caption{Diagram of the global attention process in the transformer, where the global attention score is performed by each token (pixel vector) pair from the entire image, resulting in a global representation of contextual associations  \cite{attentionisallyoueed}.}
    \label{fig:selfattention}
\end{figure*}
%

First, an encoder with two convolutional layers is employed to generate feature maps without haze components from the {clean/informative} MSI.
{As illustrated in Figure \ref{fig:framework} (cf. SR block)}, by the SR encoder (i.e., $\text{Encoder}(\cdot)$), we indeed obtain ten high-quality low-level feature maps denoted as  $\bS\in \mathbb{R}^{H\times W \times 10}$, which preserves the {valuable} spatial information extracted from original {haze-corrupted} HSI.
After that, the SR decoder {(i.e., $\text{Decoder}(\cdot)$)} serves the purpose of acquiring non-linear mappings from high-quality feature maps to the target clean/informative HSI {that includes complete 172 spectral bands.}
In summary, the SR encoder derives the feature maps $\bS$ from the {clean/informative} MSI $\bY_s$ and the SR decoder reconstructs the  spectrally {super-resolved} HSI $\bX_{\text{pre}}$, defined as follows:
\begin{align}
\bS&= \text{Encoder}(\bY_s),
\\
\bX_{\text{pre}}&= f_{\theta}(\text{Decoder}(\bS)\copyright\bY),
\end{align}
where $\copyright$ is the concatenation operator; 
$f_{\theta}(\cdot)$ stands for the $3\times3$ convolutional layer; $\theta$ means the corresponding learning weights in $f_{\theta}(\cdot)$.
{Instead of directly adopting skip connections, we merge two inputs by performing a linear projection with flexible and learnable weights to achieve the same objective.}
The effectiveness of the proposed SR block will be demonstrated in Section \ref{sec:experiment}.
Moreover, it is worth noting that the initial output obtained from the SR block is affected by color distortion and spectral angle error. 
Based on this observation, an additional refinement block that enhances the details of the spectral and spatial dimensions of the spectrally {super-resolved} solution is necessary, and this task will be detailed in Section \ref{sec:ssr}.
%
%

%
%
\subsection{Spectral-Spatial Enhancement (SSE)}\label{sec:ssr}
%
%
%
The refinement step aims at enhancing the spectral and spatial details to mitigate color distortion and signature distortion. 
In this subsection, we present the final block of our proposed {deep transformer} network, which employs the transformer structure to further enhance the details of the resulting image.
Compared to the CNN layer that is limited to a local receptive field \cite{wang2020cnn}, the transformer structure provides superior flexibility in handling longer input sequences and considering long-range dependencies by a global self-attention mechanism (cf. Figure \ref{fig:selfattention}).
However, {for} HSI data, both spectral details and spatial details are equally crucial, where the former is for material identification while the latter is for material classification.
Therefore, a joint spectral and spatial transformer structure is adopted in {the SSE} block to perform three-dimensional restoration on the HSI. 
%
%
The {proposed deep transformer} is the first one that introduces the spatial-spectral transformer into the HyDHZ problem. 

{To facilitate comprehension of the transformer \cite{attentionisallyoueed}, Figure \ref{fig:selfattention} illustrates a key process of the global attention mechanism $\text{Att}_\text{spa}(\cdot)$ in the spatial dimension by multiplying the global attention score $\bG_\text{spa} \in \mathbb{R}^{HW\times HW}$, defined as follows:}
%
%
\begin{equation}
\text{Att}_\text{spa}(\mathbf{X}) \equiv {\text{ Attention}_\text{spa}}(\bQ, \bK, \bV)=\bG_\text{spa}\bV,
\end{equation}
where $\bQ$, $\bK$, and $\bV$ represent the linear projected matrices of query, key, and value obtained by the initial 2-D flattened token $\bX$ $\in \mathbb{R}^{HW \times C}$ from 3-D tensor $\mathbf{X}\in \mathbb{R}^{H\times W \times C}$, respectively;  $\bG_\text{spa}={\text{ softmax}}\left ({\bQ\bK^{T}/\sqrt {\bd_k}}\right)$; $\bd_k$ means the scaling value.
%
%
%
%
%
Specifically, the SSE block is designed to alternatively conduct both spectral and spatial attention using transformer blocks with global self-attention, which allows the SSE block to extract the necessary spectral and spatial information required for refining the spectrally super-resolved solution.
$\text{Att}_\text{spe}(\mathbf{X})$ is another variant attention mechanism that works along the spectral
 dimensions, defined as follows:
\begin{align}
\text{Att}_\text{spe}(\mathbf{X}) &\equiv {\text{ Attention}_\text{spe}}(\bQ, \bK, \bV) =\bV \bG_\text{spe},
\end{align}
where $\bG_\text{spe}={\text{ softmax}}\left ({\bK^{T}\bQ}/\sqrt {\bd_k}\right)\in \mathbb{R}^{C\times C }$. 
After that,  the feed-forward network (FFN) is exploited to extract higher-order features from the global self-attention outcome. 
%

%
It is noteworthy that the refinement block can be modulated by the attention mechanism employed in either spectral or spatial dimensions (i.e., $\text{Att}_\text{spe}(\cdot)$ and $\text{Att}_\text{spa}(\cdot)$), which are defined as spectral refinement (i.e., SpeR($\cdot$)) and spatial refinement (i.e., SpaR($\cdot$)), respectively.
The corresponding procedure of the spatial refinement block $\text{SpaR}(\cdot)$ and spectral refinement block $\text{SpeR}(\cdot)$ can be mathematically defined as follows:
\begin{align}
\text{SpaR}(\mathbf{X})&= \text{FFN}(\text{Att}_\text{spa}(\mathbf{X})+\mathbf{X})+(\text{Att}_\text{spa}(\mathbf{X})+\mathbf{X}),
\\
\text{SpeR}(\mathbf{X})&= \text{FFN}(\text{Att}_\text{spe}(\mathbf{X})+\mathbf{X})+(\text{Att}_\text{spe}(\mathbf{X})+\mathbf{X}),
\\
\text{FFN}(\mathbf{X})&=f_{\theta_3}(\sigma(f_{\theta_2}(\sigma(f_{\theta_1}(\mathbf{X}))))),
\end{align}
where $f_{\theta_i}(\cdot)$ is the $i$th convolutional layers of the FFN block, as illustrated in Figure \ref{fig:framework}; $\theta_i$ means the corresponding learning weights in $f_{\theta_i}(\cdot)$.
As shown in Figure \ref{fig:framework}, the SSE block $\text{SSE}(\cdot)$ is composed of alternating spectral 
 and spatial transformer blocks, which is explicitly defined as follows:
\begin{align}
    \text{SSE}(\mathbf{X}) &= f_{\theta}(\text{SpaR}(\text{SpeR}(\text{SpaR}(\text{SpeR}(\mathbf{X})))))+\mathbf{X},
\end{align}
where $f_{\theta}(\cdot)$ is the $3\times3$ convolutional layer, as illustrated in Figure \ref{fig:framework}.
Furthermore, the effectiveness of the proposed SSE block will be demonstrated in Section \ref{sec:ablation}.

\subsection{Optimization of T$^2$HyDHZ}\label{sec:loss}

Traditional loss functions like L1 loss \cite{DCSN} and mean squared error (MSE) loss \cite{guo2020rsdehazenet} may not be suitable for training HSIs, as they fail to take into account the impact caused by different brightness levels of bands in HSI data.
Apparently, the luminance difference over elements will result in recovering bias since darker areas have less impact on the overall loss {calculation}.
Therefore, when designing the loss function for HSI data, it is crucial to consider the differences in luminance levels across different channels. 
To address this issue, the mean relative absolute error (MRAE) \cite{mraetgrs} is adopted in designing the loss function. 

The MRAE is formulated as follows:
\begin{align*} \text {MRAE}=&\frac {1}{N}\sum _{i=1}^{N} \frac{{|{\bX}_i-{\widehat{{\bX}_i}}|}}{{\bX}_i},
\end{align*}
where $\bX_i$ and ${\widehat{{\bX}_i}}$ denote the $i$th pixel in the ground truth (GT) and the reconstructed HSI, respectively. $N$ represents the total number of pixels.
In order to avoid unstable convergence caused by a small denominator, the refined MRAE (rMRAE) \cite{shi2018deep} is presented as follows: 
\begin{align*} \text {rMRAE}=&\frac {1}{N}\sum _{i=1}^{N} \frac{|{\bX}_i-{\widehat{{\bX}_i}}|}{{\bX}_i+1},
\end{align*}
which is adopted to pointwisely measure the difference between the predicted image and the GT image.
%
%

Considering the fact that infrared bands are less affected by haze compared to visible bands, an L1 sparsity loss $L_\text{sparsity}(\cdot)$ is
applied to promote the sparsity of visible bands from the ABS block's output.
This loss aims to reduce the likelihood of selecting low-quality visible bands, which can be mathematically described as follows:
\begin{align*}L_\text{sparsity}= \vert\vert[\bY_s]_\text{visible}\vert\vert_{1},
\end{align*}
where $[\cdot]_\text{visible}$ denotes the specific visible bands.
All in all, by incorporating the rMRAE loss and the L1 sparsity loss, the adopted training loss can be presented as follows:
\begin{align}\label{eq:Ldef}
{L}= {L}_\text{rMRAE}+ {L}_\text{sparsity},
\end{align}
where ${L}_\text{rMRAE}$ indicates the rMRAE loss; ${L}_\text{sparsity}$ means the L1 sparsity loss promoted over visible bands for obtaining {clean/informative MSI} via the ABS block. 
Then, \eqref{eq:Ldef} is optimized to train the proposed T$^2$HyDHZ via the Adam
optimizer \cite{kingma2014adam} with adaptive learning rate strategy.
{Furthermore, we have conducted an ablation study \textit{with/without} $L_\text{sparsity}$ loss to assess its impact on quantitative performance, as summarized in Section \ref{sec:ablation}.}
%
%

\section{Experimental Results}\label{sec:experiment}
In this section, we will provide details on the training process of the proposed T$^2$HyDHZ method and present {its} performance through qualitative and quantitative evaluations.
Section \ref{sec:experimentsetting} introduces the dataset and the network hyperparameters.
Sections \ref{sec: Qualitative Results} and \ref{sec:quantitativeassessment} discuss the qualitative and quantitative assessment results, respectively.
Section \ref{sec:sythesizedata} illustrates the generating process of synthesized hazy datasets for network training and testing.
{In Section \ref{sec:ablation}, we conduct ablation studies to evaluate the contributions of individual blocks systematically, to assess the efficacy of the $L_\text{sparsity}$ loss, and to evaluate the benefits of concatenation component within the framework.}
%

\subsection{Experimental Setting}\label{sec:experimentsetting}

\textit{Dataset Description:}
In this paper, a dataset consisting of  2.8K HSIs acquired from the AVIRIS sensor \cite{portal} is employed. 
The HSIs in this dataset have a dimension of 256 $\times$ 256 $\times$ 172\cite{DCSN}, which covers a wavelength range of 0.4-2.5 $\mu$m and encompasses a variety of landscapes, including farms, cities, mountains, and coastal regions across the USA and Canada. 
In order to ensure the image quality of the collected HSI for training and testing, we further eliminate low-quality bands severely affected by water vapor, including bands 1–10, 104–116, 152–170, and 215–224, as reported in \cite{LinTGRS2021}.
Additionally, to maximize the data utilization for training the model, we randomly allocate 90\% of the data for training, 5\% for validation, and the remaining 5\% for testing.
%
\\

\textit{Network Hyperparameters:}
Here outlines the training procedure of the proposed model, which involves using Adam optimizer \cite{kingma2014adam} with a learning rate decay strategy. 
Specifically, the initial learning rate is set to $3\times10^{-4}$, with a decay rate of 0.6 after every 30 epochs. 
The termination criterion for the training process is the convergence of the validation loss or, at most, 300 epochs, which indicates that the model has attained an acceptable level of performance.
Besides, the computational setup and resources are summarized below.
The training phase is carried out on a desktop computer equipped with an NVIDIA RTX 3090 GPU and an Intel Core i9-10900K CPU (3.70 GHz speed and 128 GB of RAM). 
Apart from the training phase, all other experiments are conducted on a separate computer equipped with an NVIDIA GTX-2080Ti GPU and an Intel Core-i7-10700K CPU (3.80 GHz speed and 32 GB of RAM). 
Furthermore, the numerical computing environment employed for the DE solutions is Python 3.6.12, while other methods are executed using Mathworks Matlab R2021b.
\subsection{Qualitative Analysis}\label{sec: Qualitative Results}
\begin{figure*}[t]
    \begin{center}
        \includegraphics[width=1\textwidth] {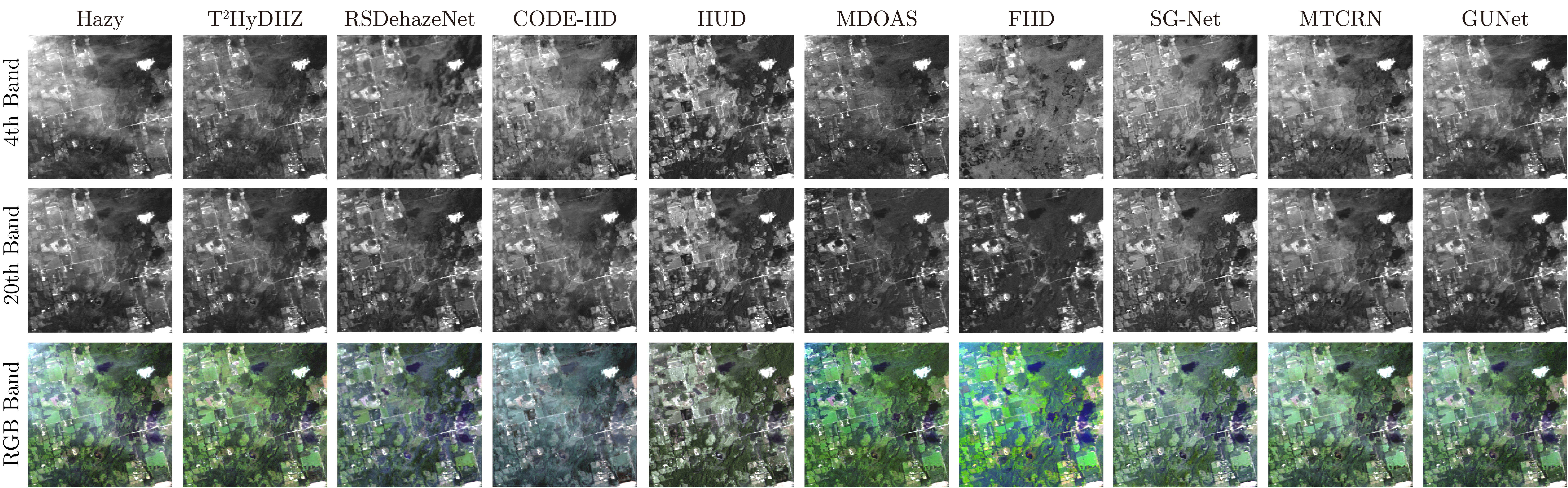}
    \end{center}
    \caption{Qualitative study of AVIRIS real hyperspectral imagery from NASA over Kettle Moraine, USA. }
    \label{fig:real1}
\end{figure*}
\begin{figure*}[t]
    \begin{center}
        \includegraphics[width=1\textwidth] {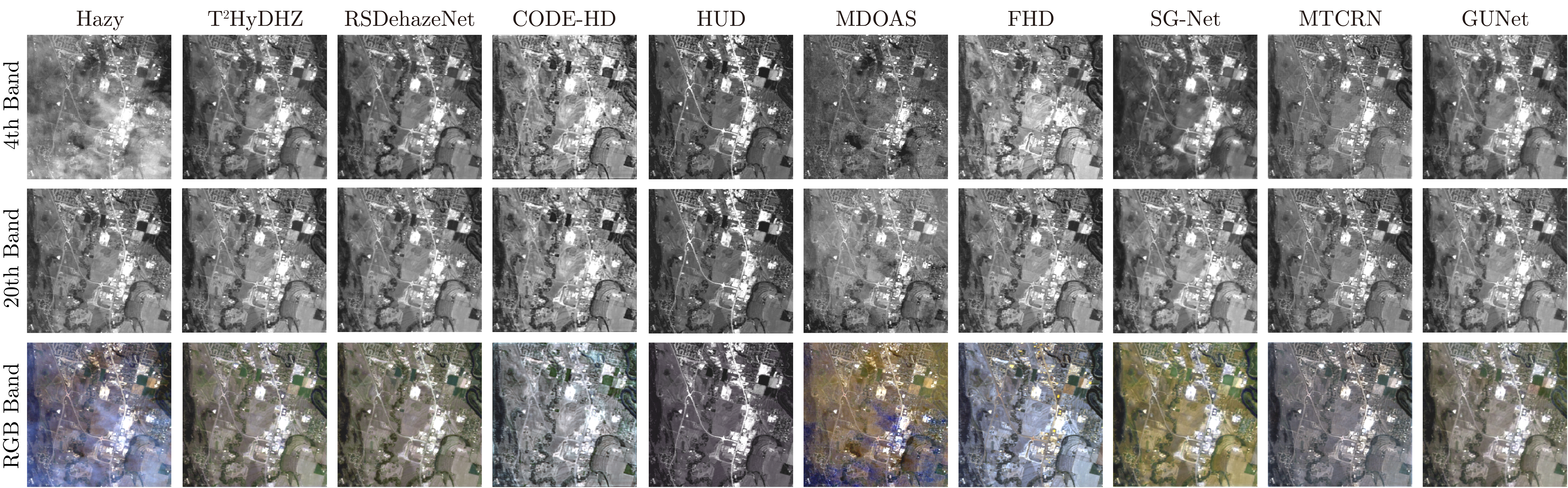}
    \end{center}
    \caption{Qualitative study of AVIRIS real hyperspectral imagery from NASA over Flathead Lake, USA.}
    \label{fig:real2}
\end{figure*}
\begin{table}[t]
\centering
\caption{Real hazy AVIRIS images used for qualitative analysis in Section 
\ref{sec: Qualitative Results}.}
\label{table:datatable}	
\setlength{\tabcolsep}{10mm}{
\tabcolsep0.05in
 \renewcommand\arraystretch{1.2}
\resizebox{\linewidth}{!}{
\begin{tabular}{ccccc}
\hline
Site Name&Type&Flight Name&Resolution&Date\\
\hline
Kettle Moraine, WI&Farm&f090729t01p00r06&16.8m&2009/7/29\\
Flathead Lake, MT&City&f190808t01p00r14&15.2m&2019/8/8\\
\hline
\end{tabular}
}
}
\end{table}
In this subsection, we compare our proposed T$^2$HyDHZ with several benchmark methods, including the NN model-based dehazing network for remote sensing (RSDehazeNet)\cite{guo2020rsdehazenet}, the CODE theory-based method (CODE-HD)\cite{Tang2022dehazing}, the hyperspectral unmixing-based dehazing method (HUD)\cite{gan2016dehazing}, the AS model-based MuDHZ method (MDOAS)\cite{GuoTGRS2021}, the fog intensity map estimation-based HyDHZ method (FHD)\cite{XudongTGRS2021}{, the spectral grouping-based HyDHZ network (SG-Net) \cite{ma2022spectral}, the UNet-guided mathematical imaging model for multispectral thin cloud removal (MTCRN) \cite{MTCRN2021jstar}, and the UNet-based remote sensing dehazing network with gating mechanisms (GUNet) \cite{song2022rethinking}.}
To evaluate the practicability of studied methods in real-world applications, we consider natural hazy AVIRIS HSIs, including two sub-scenes acquired by the AVIRIS sensor over Kettle Moraine and Flathead Lake regions, which are across the farm and urban sub-scenes with the size of $256\times256\times224$.
However, considering potential pollution in the original 224 bands of the AVIRIS data, our network is trained solely on 172 high-quality bands, excluding bands 1-10, 104-116, 152-170, and 215-224, as reported in \cite{LinTGRS2021}.
Thus, we only evaluate the qualitative performance of the real hazy data using the remaining 172 bands in this subsection.
Furthermore, additional details regarding the acquisition of the real hazy data can be found in Table \ref{table:datatable}, where interested readers may access the data via the AVIRIS data portal{\footnote{https://aviris.jpl.nasa.gov/dataportal/.}} with sufficient information.

Figures \ref{fig:real1} and \ref{fig:real2} illustrate two hazy scenarios over farm and city types investigated for qualitative analysis among the studied methods, displayed in RGB bands 19, 9, and 2.
It can be observed that although RSDehazeNet is competitive, the proposed T$^2$HyDHZ method exhibits less color distortion due to its spectral-spatial refinement mechanism, which effectively preserves both spectral and spatial information and minimizes the bias of the pixel value of the recovered image. 
On the other hand, CODE-HD, HUD, MDOAS, FHD, {SG-Net, MTCRN, and GUNet} suffer from varying degrees of color distortion because of their inability to eliminate the scattered light effect that causes pollution and to recover the original colors and details of the HSI.
For example, as illustrated in Figure \ref{fig:real1}, all existing peer methods do not reserve correct color details even in the non-hazy region, resulting in varying degrees of color distortion overall.
By contrast, the proposed T$^2$HyDHZ method outperforms benchmark methods in the qualitative analysis {using} real hazy data acquired by the AVIRIS sensor, showing less color aberration in the recovered image.
Besides, the effectiveness of T$^2$HyDHZ will be further confirmed through the quantitative analysis in Section \ref{sec:quantitativeassessment}, where its superior performance will be compared with benchmark peer methods with more in-depth examinations.
\begin{figure}[t]
    \begin{center}      \includegraphics[width=0.49\textwidth] {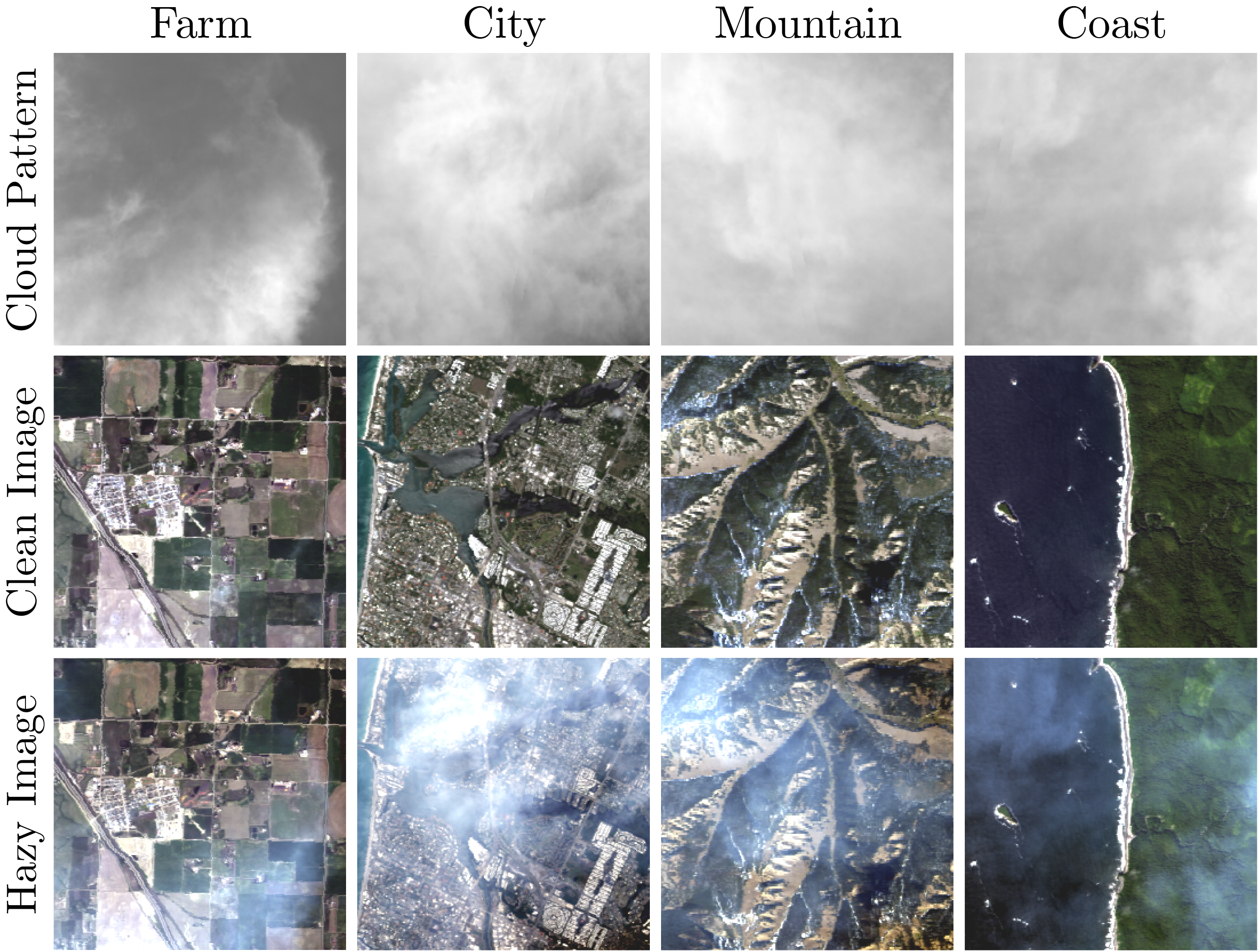}
    \end{center}
    \caption{
%
%
%
Synthesized hazy HSI using authentic, clean images and actual cloud patterns. 
The first row shows four patterns extracted from the cirrus band ($\bB_9$) of Landsat-8 OLI data over the region of Merauke, Papua New Guinea (Scene ID: LC81000652022332LGN00, acquired on November 28, 2022).  
The second row shows the clean images of four natural scenes for the subsequent haze simulating, while the last row shows the simulated hazy HSIs.}
    \label{fig:gernate haze}
\end{figure}
\subsection{Preparation of Simulated Hazy Data}\label{sec:sythesizedata}

%
%
For data-driven methods, one of the most critical processes is the preparation of training HSI data pairs, involving the collection of clean and hazy datasets over the same region at different times, which can be quite time-consuming and impractical.
To conquer this challenge, a haze simulation strategy is adopted in this study for generating training data pairs more efficiently and practically, as reported in recent studies \cite{guo2020rsdehazenet,ma2022spectral}. 
Therefore, it enables us to create diverse training data pairs with different haze conditions (cf. Figure \ref{fig:gernate haze}) that appropriately represent various hazy circumstances in real-world scenarios without requiring extensive data collection efforts (i.e., collecting the hazy and non-hazy data in the same region).
Moreover, considering various scenarios by this haze simulation strategy allows us to construct various intensities of haze levels while generating training data pairs under different haze conditions to help prevent overfitting to specific hazy patterns and enhance the robustness of the model.
%
%
%

To synthesize the simulated hazy data, we initially acquire the cirrus band $\bB_9$$\in \mathbb{R}^{H \times W}$ of the Landsat-8 OLI from the United States Geological Survey (USGS)\footnote{https://earthexplorer.usgs.gov/}.
Afterward, we randomly crop several haze/cloud-covered patches $\bb_9$ $\in \mathbb{R}^{h \times w}$ over multiple regions of interest (ROI), as demonstrated in the first {row} of Figure \ref{fig:gernate haze} ($h$=$w$=256 in this work). 
Next, we calculate the reference transmission map $\bt_1\in \mathbb{R}^{h \times w}$ and the transmission map of each band $\bt_c \in \mathbb{R}^{h \times w}$ using the following definitions:
\begin{align}
\label{eq:t1}
 &\bt_1(x)=1-\alpha \bb_9(x), 
 \\
  \label{eq:t2}&\bt_c(x)\:=e^{(\frac{\lambda_1}{\lambda_c})^{\gamma(x)}\text{ln}\bt_1(x)},
 \end{align}
 where $\bb_{9}\in \mathbb{R}^{h \times w}$ denotes the cropped ROI of  $\bB_9$ in Landsat-8 OLI data; $\alpha$ is the control constant of different hazy levels; $\lambda_c$ is the central wavelength of band $c$; $\gamma(x)$ is the pixel-wise parameter of the haze concentration; $x$ is the spatial location.
%
Subsequently, the corresponding transmission maps can be obtained by \eqref{eq:t1} and \eqref{eq:t2}.
Finally, we can swiftly generate simulated hazy data with collected GT data (see the second and third {rows} in Figure \ref{fig:gernate haze}) through the AS model \cite{mccartney1976optics,chavez1988improved} as follows:
\begin{align}
\label{eq:asmodel}
 \bH_{c}(x)=\bG_{c}(x)\bt_c(x)+\bA_{c}(x)\big(1-\bt_c(x)\big),
 \end{align}
 where $\bH_c$, $\bG_c$, and $\bA_c$ $\in \mathbb{R}^{h \times w}$ denote the simulated hazy image, the GT image, and global atmospheric light of band $c$, respectively. 

\begin{figure}[t]
	\begin{minipage}{0.324\linewidth}
		\centerline{\includegraphics[width=1\textwidth]{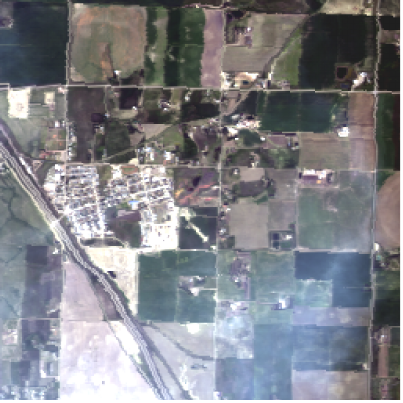}}
		\vspace{-0.16cm}
		\caption*{(a) $\alpha = 0.5$}
        \vspace{0.2cm}
	\end{minipage}
	\begin{minipage}{0.324\linewidth}
		\centerline{\includegraphics[width=1\textwidth]{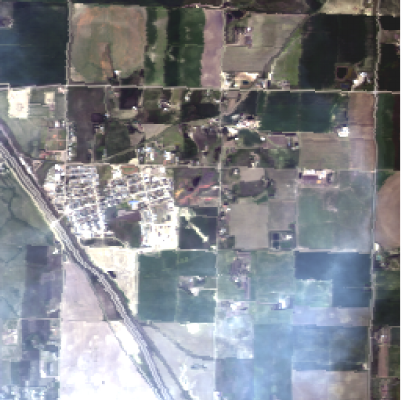}}
		\vspace{-0.16cm}
		\caption*{(b) $\alpha = 0.6$}
        \vspace{0.2cm}
	\end{minipage}
        \begin{minipage}{0.324\linewidth}
	\centerline{\includegraphics[width=1\textwidth]{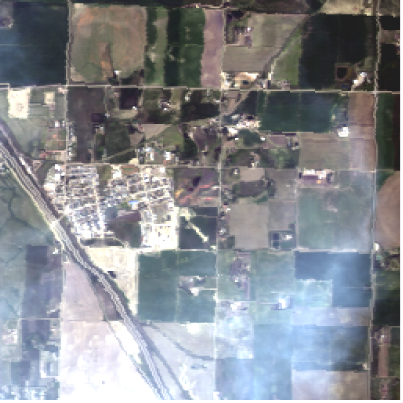}}
		\vspace{-0.16cm}
		\caption*{(c) $\alpha = 0.7$}
        \vspace{0.2cm}
	\end{minipage}
 \begin{minipage}{0.324\linewidth}
		\centerline{\includegraphics[width=1\textwidth]{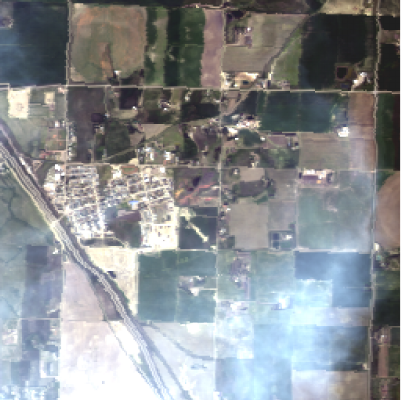}}
		\vspace{-0.15cm}
		\caption*{(d) $\alpha = 0.8$}
	\end{minipage}
	\begin{minipage}{0.324\linewidth}
		\centerline{\includegraphics[width=1\textwidth]{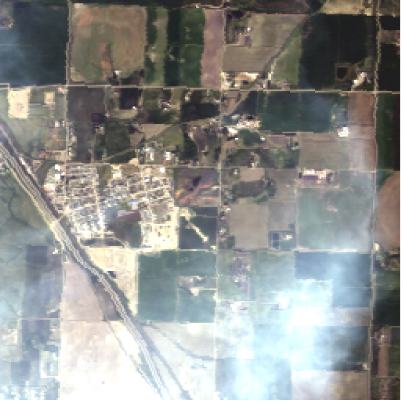}}
		\vspace{-0.15cm}
		\caption*{(e) $\alpha = 0.9$}
	\end{minipage}
        \begin{minipage}{0.324\linewidth}
	\centerline{\includegraphics[width=1\textwidth]{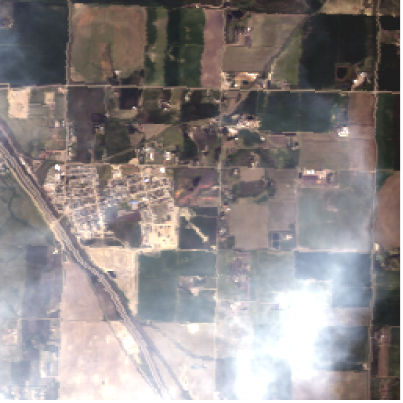}}
		\vspace{-0.15cm}
		\caption*{(f) $\alpha = 1$}
  
	\end{minipage}
        
	\caption{Diverse levels of the haziness of synthesized hazy images under various $\alpha$ conditions, where $\alpha$ exhibits a direct correlation with the degree of atmospheric obscurity.}\label{fig:omega0.8-1}
\end{figure}

In order to generate a training dataset that accurately resembles real-world scenarios with varying haze intensity levels (cf. Figure \ref{fig:omega0.8-1}), we set $\gamma(x):=3$ and $\alpha\in$ $\{0.5, 0.6, 0.7, 0.8, 0.9, 1\} $ to synthesize data that conforms to the real haze distribution on bands. The haze distribution experience is from the observation on the real hazy AVIRIS dataset and the previous study \cite{XudongTGRS2021}.  
 For calculating $\bA_c$ in \eqref{eq:asmodel}, we average the 0.01\% brightest pixels by bands and share the value among all pixels within each band $c$ as stated in \cite{hetpami2010,guo2020rsdehazenet}. 
In addition, we adopt data augmentation techniques, including degree rotations and random horizontal/vertical flips to both GT image $\bG \in \mathbb{R}^{h \times w \times c}$ and transmission maps $\bt \in \mathbb{R}^{h \times w \times c}$, which is implemented before simulating haze pollution on HSIs.
Furthermore, we allocate 80\% of the generated hazy patterns for training, while the remaining patterns are equally reserved for validation and testing purposes  (i.e., various hazy patterns obtained from \eqref{eq:t1} to \eqref{eq:asmodel} with different ROI).
%

%
\subsection{Quantitative Assessment}\label{sec:quantitativeassessment}
In this subsection, four commonly used objective metrics for dehazing evaluations are adopted to assess the performance of the proposed T$^2$HyDHZ algorithm with benchmark MuDHZ/HyDHZ methods, including peak signal-to-noise
ratio (PSNR)\cite{psnr}, universal image quality index (UIQI)\cite{UIQI2002}, spectral angle mapper (SAM)\cite{SAM92,lin2017COCNMF}, and structural similarity (SSIM)\cite{ssim}.
All quality metrics in this subsection are calculated using the reference image $\bX$ $\in \mathbb{R}^{H\times W \times C}$ (i.e.,  the clean GT) and the reconstructed image $\widehat\bX$ $\in \mathbb{R}^{H\times W \times C}$, while their definitions are then provided in the following paragraphs, respectively.

\begin{figure*}[t]
    \begin{center}
        \includegraphics[width=1\textwidth] {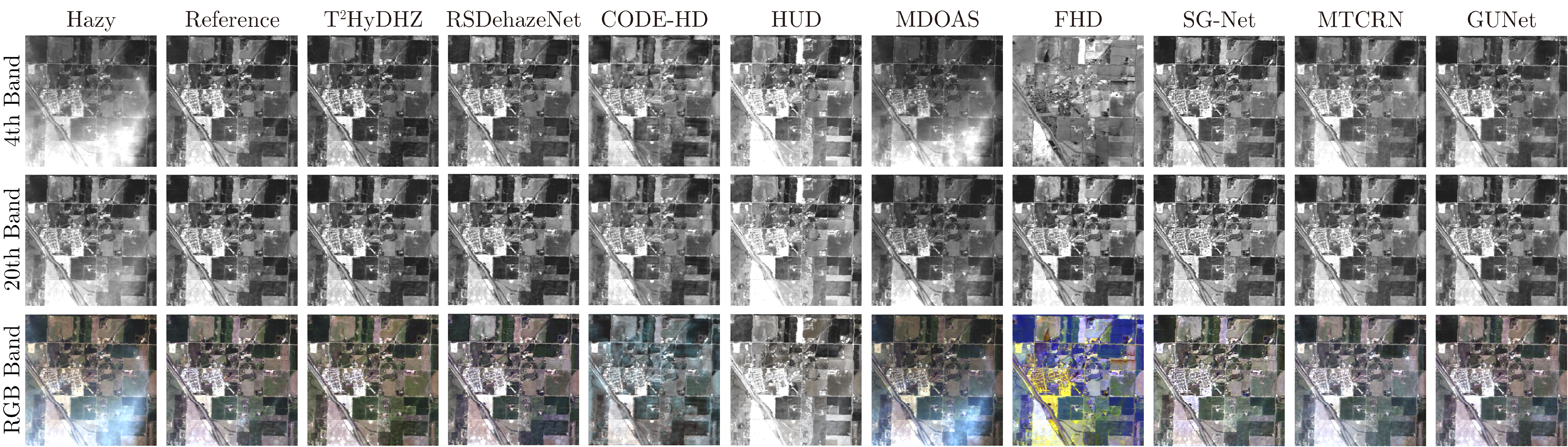}
    \end{center}
    \caption{Dehazing results of the simulated hazy hyperspectral imagery over Alberta, Canada.}
    \label{fig:exp1}
\end{figure*}
\begin{figure*}[t]
    \begin{center}
        \includegraphics[width=1\textwidth] {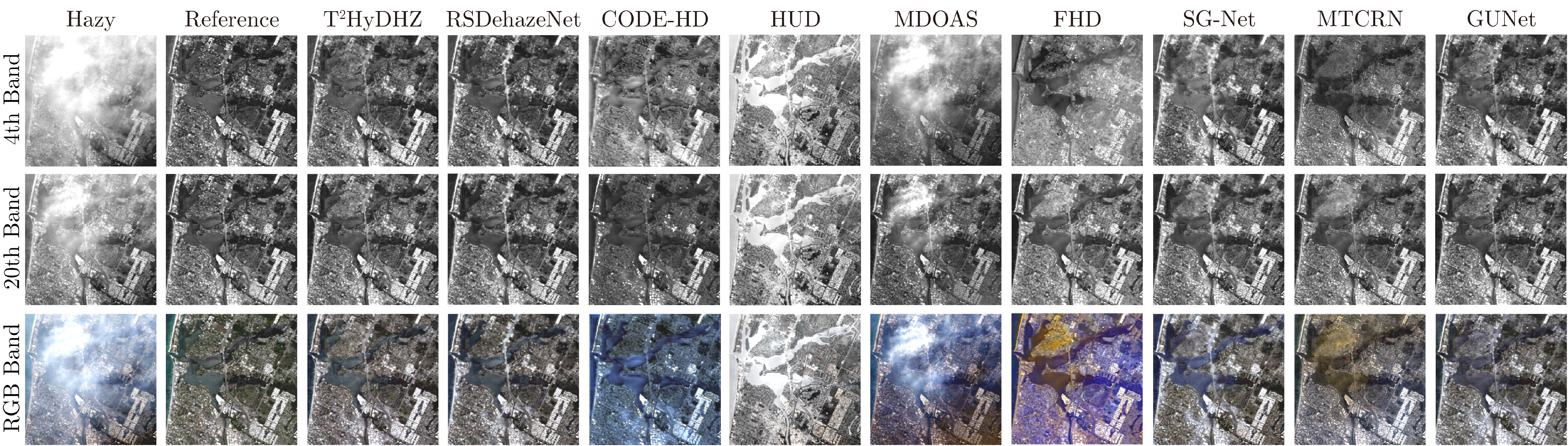}
    \end{center}
    \caption{Dehazing results of the simulated hazy hyperspectral imagery over Florida, USA.}
    \label{fig:exp2}
\end{figure*}
%
%
    \begin{table}[t]
	\centering
	\caption{Quantitative performance assessment of various dehazing
methods using the Alberta data.}\label{table:peertable1}
	\vspace{-0.1cm}
  \renewcommand\arraystretch{1.1}
 \scalebox{1.05}{
	\setlength{\tabcolsep}{1.2mm}{

		\begin{tabular}{c|ccccc}
			\hline
		Methods&PSNR~ & UIQI~ &SAM~&SSIM~ &Time (sec.)
			\\
			\hline
		T$^2$HyDHZ&\bf 46.102&\bf 0.996&\bf 1.062&\bf 0.999&\bf 0.008
			\\
        RSDehazeNet\cite{guo2020rsdehazenet}&37.113&0.978&1.825&0.991&0.061
		  
			\\	
        CODE-HD\cite{Tang2022dehazing}&30.722&0.942&4.807&0.968&186 
			\\
       HUD\cite{gan2016dehazing}&16.382&0.623&11.373&0.794& 4.244
	   \\
        MDOAS\cite{GuoTGRS2021}&21.228&0.851&11.199&0.926&2.608
			\\	
        FHD\cite{XudongTGRS2021}&25.232&0.861&24.851&0.865& 0.136
			\\

        SG-Net\cite{ma2022spectral}& 34.925& 0.961& 2.678& 0.984& 0.016
			\\

        MTCRN\cite{MTCRN2021jstar}& 39.689& 0.993& 2.072& 0.996& 0.009
			\\

        GUNet\cite{song2022rethinking}& 44.255& 0.994& 1.536& 0.998& 0.062
			\\

			\hline
		\end{tabular}}}
\end{table}
%
    \begin{table}[t]
	\centering
	\caption{Quantitative performance assessment of various {dehazing}
methods using the Florida data.}\label{table:peertable2}
	\vspace{-0.1cm}
 \renewcommand\arraystretch{1.1}
 \scalebox{1.05}{
	\setlength{\tabcolsep}{1.2mm}{
		\begin{tabular}{c|ccccc}
			\hline
		Methods&PSNR~ & UIQI~ &SAM~&SSIM~ &Time (sec.)
			\\
			\hline
		T$^2$HyDHZ&\bf 35.342&\bf 0.981&\bf 2.594&\bf 0.988&\bf0.007
			\\
        RSDehazeNet\cite{guo2020rsdehazenet}&30.475&0.931&4.37&0.964&0.022
		  
			\\	
        CODE-HD\cite{Tang2022dehazing}&28.109&0.919&6.07&0.933&187.577 
			\\
       HUD\cite{gan2016dehazing}&18.434&0.573&19.363&0.732&4.416
	   \\
        MDOAS\cite{GuoTGRS2021}&10.332&0.403&28.042&0.477&2.148
			\\	
        FHD\cite{XudongTGRS2021}&20.798&0.452&70.904&0.503&0.126
			\\

        SG-Net\cite{ma2022spectral}& 30.868& 0.927& 4.295& 0.965& 0.014
			\\

        MTCRN\cite{MTCRN2021jstar}& 30.233& 0.934& 9.798& 0.948& 0.011
			\\

        GUNet\cite{song2022rethinking}& 33.685& 0.951& 3.553& 0.970& 0.056
			\\

			\hline
		\end{tabular}}}
\end{table}

The PSNR is to evaluate the spatial reconstructed quality between the reference image $\bX$ and the recovered image $\widehat\bX$, defined as follows:
\begin{equation*} 
\text {PSNR}=\frac {1}{C}\sum _{i=1}^{C}
10\log _{10}\left ({\frac {\max \big (\bX_{i}\big)^{2}}{\frac {1}{L}\| \bX_i-\widehat{\bX}_i\|_{F}^{2}}}\right ), 
\end{equation*}
where $C$ represents the number of spectral bands. $\bX_i$ and $\widehat{\bX}_i$ are the $i$th band of the reference image $\bX$ and the reconstructed image $\widehat{\bX}$, respectively. $L$=$H$$W$ is the total number of pixels in each band.

The UIQI index is to evaluate the loss of correlation, luminance distortion, and contrast distortion between the reference image $\bX$ and the recovered image $\widehat\bX$, defined as follows: 
\begin{equation*} 
\text {UIQI}= \frac {1}{C}\sum _{i=1}^{C} \mathbb {E}\left [{ \frac {\sigma _{\bx_i\widehat{ {\bx}}_i}}{\sigma _{\bx_i}\sigma _{\widehat{ {\bx}}_i}} \cdot \frac {2~\mu _{\bx_i}\mu _{\widehat {\bx}_i}}{\mu _{\bx_i}^{2}+\mu _{\widehat {\bx}_i}^{2}} \cdot \frac {2\sigma _{\bx_i}\sigma _{\widehat {\bx}_i}}{\sigma _{\bx_i}^{2}+\sigma _{\widehat {\bx}_i}^{2}} }\right],
\end{equation*}
where $\mathbb {E}$ is an averaging operator over sliding windows set as 64 $\times$ 64; $\bx_i$ is the GT image $\bX_i$ across the sliding windows and $\widehat\bx_i$ is its estimated; $\sigma _{\bx_i\widehat {\bx}_i}$ denotes the covariance between $\bx_i$ and $\widehat{\bx}_i$; $\sigma _{\bx_i}$ and $\sigma _{\widehat {\bx}_i}$ are the standard deviations of $\bx_i$ and $\widehat {\bx}_i$; $\mu _{\bx_i}$ and $\mu _{\widehat {\bx}_i}$ represent for the means of $\bx_i$ and $\widehat {\bx}_i$, respectively.

The SAM index compares the similarity in spectral characteristics between the reference image $\bX$ and the recovered image $\widehat\bX$, defined as follows:
\begin{equation*} \text {SAM}= \frac {1}{L} \sum _{i=1}^{L} \text {arccos}\left ({\frac {\by_{i}^{T}{\widehat {\by}_{i}}}{ \|{\by}_{i}\|_{2}\cdot \|{\widehat {\by}_{i}}\|_{2}}}\right),
\end{equation*}
noting that ${\bY\triangleq[\by_1,\dots,\by_L]}$ and ${\widehat{\bY}\triangleq[\widehat{\by}_1,\dots,\widehat{\by}_L]}$ $\in \mathbb{R}^{C\times L}$ are 2-D matrices being reshaped from 3-D tensors $\bX$ and $\widehat{\bX}$.

The SSIM index can be regarded as an extended version of UIQI defined as follows:
\begin{align*} 
&\text {SSIM}_i \notag
\\
&=\mathbb {E}\left [{ \frac {2\mu _{ \bx_i}\mu _{\widehat { \bx}_i}+C_{1}}{\mu _{ \bx_i}^{2}+\mu _{\widehat { \bx}_i}^{2}+C_{1}} \cdot \frac {2\sigma _{ \bx_i}\sigma _{\widehat { \bx}_i}+C_{2}}{\sigma _{ \bx_i}^{2}+\sigma _{\widehat { \bx}_i}^{2}+C_{2}} \cdot \frac {\sigma _{ \bx_i\widehat { \bx}_i}+C_{3}}{\sigma _{ \bx_i}\sigma _{\widehat { \bx}_i}+C_{3}} }\right], 
\end{align*}
%
where small constants $C_1, C_2$, and $C_3$ are set to avoid instability when the denominator is very close to zero. In this paper, $\text {SSIM} = \frac {1}{C}\sum _{i=1}^{C} \text {SSIM}_i$ with sliding windows set as 11 $\times$ 11.
\begin{figure*}[t]
    \begin{center}
        \includegraphics[width=1\textwidth] {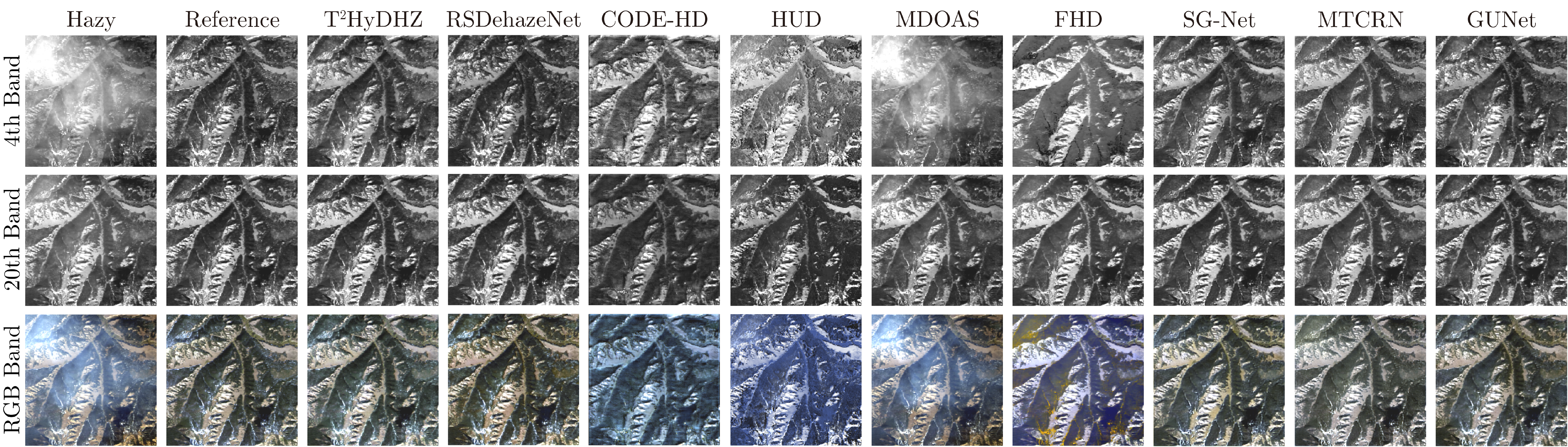}
    \end{center}
    \caption{Dehazing results of the simulated hazy hyperspectral imagery over Yellowstone National Park, USA.}
    \label{fig:exp3}
\end{figure*}
\begin{figure*}[t]
    \begin{center}
        \includegraphics[width=1\textwidth] {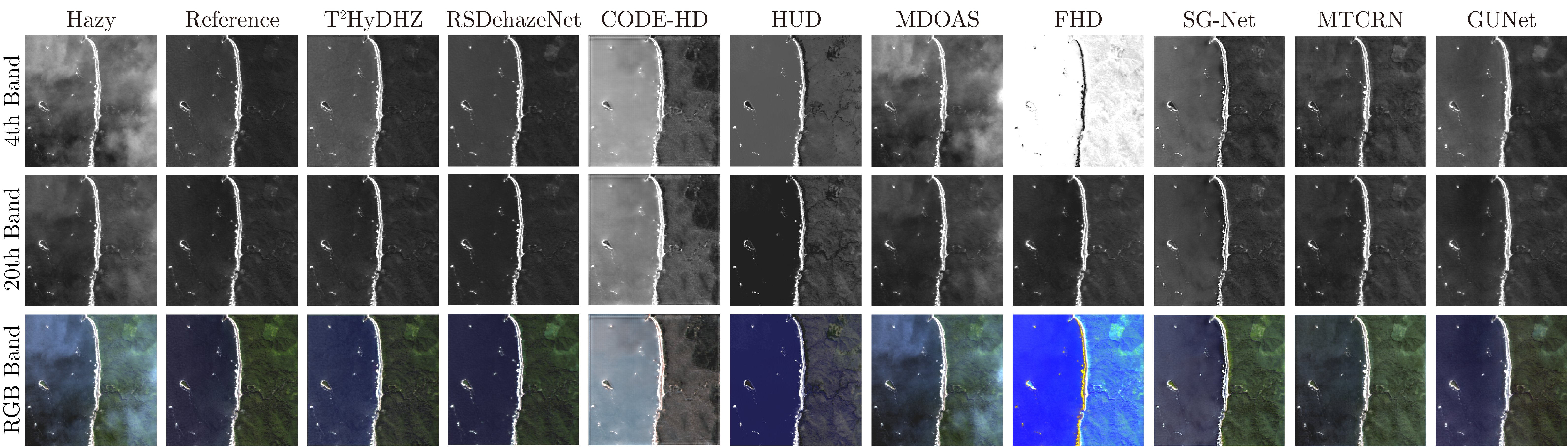}
    \end{center}
    \caption{Dehazing results of the simulated hazy hyperspectral imagery over Olympic National Park, USA.}
    \label{fig:exp4}
\end{figure*}
%
    \begin{table}[t]
    
	\centering
	\caption{Quantitative performance assessment of various dehazing
methods using the Yellowstone National Park data.}\label{table:peertable3}
	\vspace{-0.1cm}
 \renewcommand\arraystretch{1.1}
 \scalebox{1.05}{
	\setlength{\tabcolsep}{1.2mm}{
		\begin{tabular}{c|ccccc}
			\hline
		Methods&PSNR~ & UIQI~ &SAM~&SSIM~ &Time (sec.)
			\\
			\hline
		T$^2$HyDHZ&\bf 43.413&\bf 0.997&\bf 1.896&\bf 0.997&\bf0.007
			\\
        RSDehazeNet\cite{guo2020rsdehazenet}&35.135&0.989&2.479&0.987&0.035  
			\\	
        CODE-HD\cite{Tang2022dehazing}&28.318&0.943&7.24&0.94&153.471 
			\\
       HUD\cite{gan2016dehazing}&16.869&0.632&25.988&0.721&4.01
	   \\
        MDOAS\cite{GuoTGRS2021}&13.587&0.738&23.46&0.552&2.86
			\\	
        FHD\cite{XudongTGRS2021}&13.036&0.628&41.596&0.572&0.115
			\\

        SG-Net\cite{ma2022spectral}&33.452&0.984&3.246&0.984&0.017
			\\

        MTCRN\cite{MTCRN2021jstar}&33.957&0.983&4.192&0.978&0.016
			\\

        GUNet\cite{song2022rethinking}&39.079&0.995&3.951&0.995&0.061
			\\
			\hline
		\end{tabular}}}
\end{table}
%
\begin{table}[t]
    
	\centering
	\caption{Quantitative performance assessment of various dehazing methods using the Olympic National Park data.}\label{table:peertable4}
	\vspace{-0.1cm}
 \renewcommand\arraystretch{1.1}
 \scalebox{1.05}{
	\setlength{\tabcolsep}{1.2mm}{
		\begin{tabular}{c|ccccc}
			\hline
		Methods&PSNR~ & UIQI~ &SAM~&SSIM~ &Time (sec.)
			\\
			\hline
		T$^2$HyDHZ&\bf 41.295&\bf 0.95&\bf 1.365&\bf 0.996&\bf0.006
			\\
        RSDehazeNet\cite{guo2020rsdehazenet}&35.333&0.797&2.022&0.977&0.025	  
			\\	
        CODE-HD\cite{Tang2022dehazing}&27.574&0.749&10.148&0.922&159.055 
			\\
       HUD\cite{gan2016dehazing}&21.142&0.513&13.028&0.849&4.043
	   \\
        MDOAS\cite{GuoTGRS2021}&12.591&0.331&9.77&0.558&2.613
			\\	
        FHD\cite{XudongTGRS2021}&14.819&0.461&50.405&0.535&0.125
			\\

        SG-Net\cite{ma2022spectral}&31.414&0.773&4.229&0.959&0.021
			\\

        MTCRN\cite{MTCRN2021jstar}&33.788&0.802&6.554&0.95&0.012
			\\

        GUNet\cite{song2022rethinking}&33.061&0.869&5.746&0.956&0.059
			\\
			\hline
		\end{tabular}}}
\end{table}
To properly evaluate the efficacy of the studied method in quantitative assessment, we consider {four} frequently encountered hazy conditions, including scenarios with light haze corruption and heavy fog, as depicted in Figure \ref{fig:gernate haze}.
Besides, the resulting quantitative assessments of {four} studied scenes are presented in {Tables \ref{table:peertable1}, \ref{table:peertable2}, \ref{table:peertable3}, and \ref{table:peertable4}} with the best performance highlighted in bold font, indicating the highest PSNR/UIQI/SSIM values, the lowest SAM value, and the fastest computational time measured in seconds (sec.).
{As one can see, the proposed T$^2$HyDHZ method significantly improves all metrics compared to baseline methods due to the SSE block, which captures global relationships in spatial and spectral dimensions for detail refinement.
By contrast, other methods lack such a mechanism to preserve spectral information, resulting in inconsistencies in spatial and spectral details reflected in PSNR and SAM values.}
%
%

For a visual illustration, Figures \ref{fig:exp1}, \ref{fig:exp2}{, \ref{fig:exp3}, and \ref{fig:exp4}} depict that all peer methods suffer from various degrees of color distortion.
Despite employing channel attention for {preserving} spectral information, the RSDehazeNet approach still manifests color distortions, especially in the right region of Figure \ref{fig:exp1}.
Even though the CODE-HD approach combines the AS model with CODE theory and utilizes the $\bQ$-norm feature extractor to distill spatial information from a rough DE solution, its effectiveness decreases under severe haze conditions {(cf. Figures \ref{fig:exp2} and \ref{fig:exp4})}.
Although the HUD method uses hyperspectral unmixing techniques to remove the haze component in its unmixing process, the haze component may be challenging to isolate, resulting in potential spatial information loss, as depicted in {Figures \ref{fig:exp1}, \ref{fig:exp2}, and \ref{fig:exp3}}.
The MDOAS method, based on the AS model, is initially designed for the recovery of multispectral images captured by Landsat, leading to considerable errors in hazy HSI reconstruction.
The FHD method estimates a fog intensity map and propagates it to each band to determine the corresponding abundances.
However, it requires selecting the same object to minimize the problem formulation as stated in \cite[Equation (7)]{XudongTGRS2021}.
Therefore, FHD decreases estimation accuracy in complicated landscapes with various objects. 
Additionally, FHD requires the assumption that hazy HSI contains a ``clean'' area to calculate fog abundance in each band, which may not be {practical} in scenarios {with either heavy or thin haze in large regions like Figures \ref{fig:exp2}, \ref{fig:exp3}, and \ref{fig:exp4}}.
In conclusion, our proposed T$^2$HyDHZ method effectively preserves information in both spectral and spatial dimensions and significantly improves performance compared to other existing benchmark dehazing methods, as evidenced by our qualitative and quantitative analysis.
\begin{figure}[t]
    \begin{center}
        \includegraphics[width=0.49\textwidth] {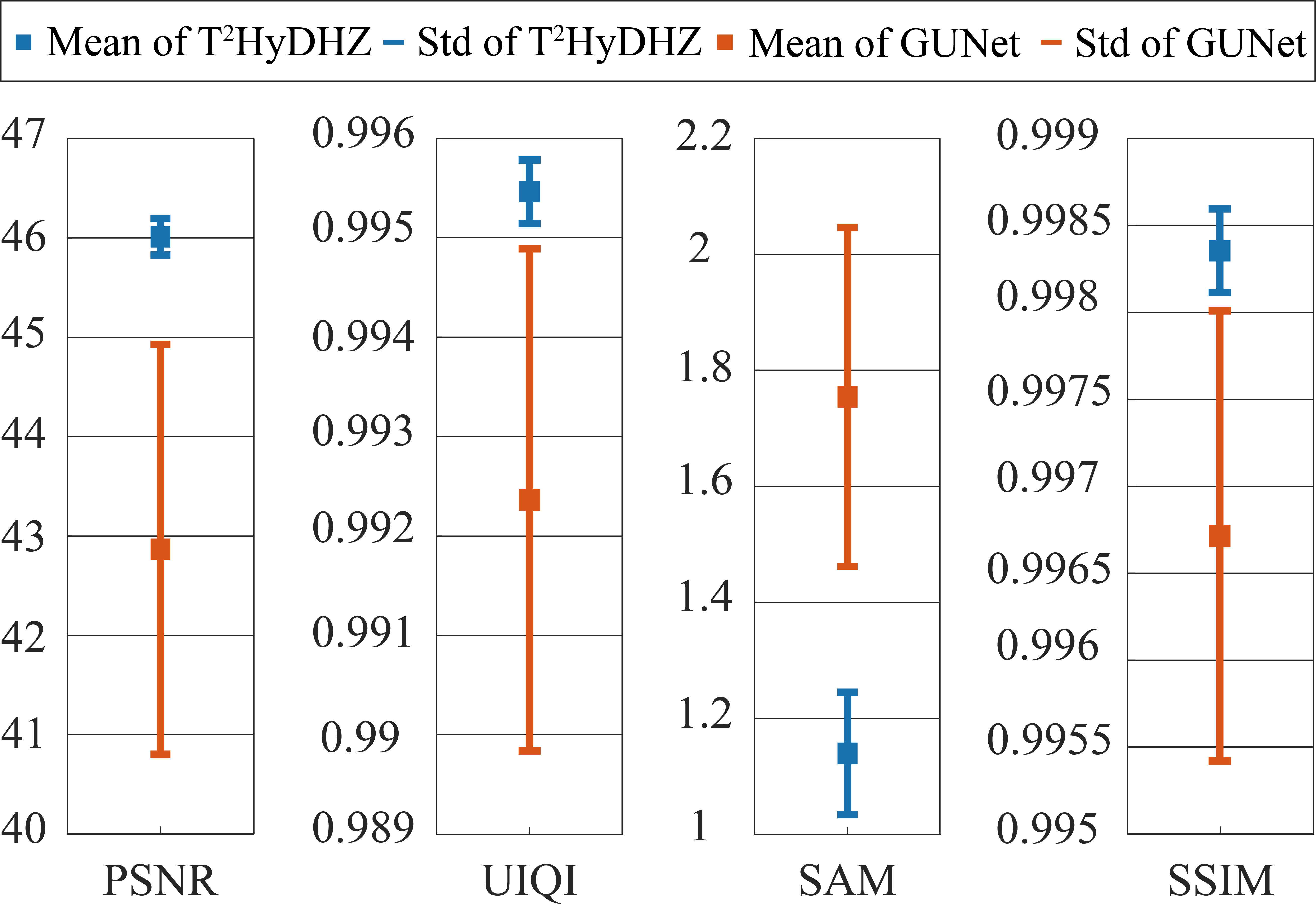}
    \end{center}
    \caption{Performance assessment of standard deviation (std) graphs tested under random haze levels compared with the state-of-the-art (SOTA) baseline (i.e., the strongest peer method).}
    \label{fig:std}
\end{figure}
{Furthermore, we have conducted a sensitivity study using random haze levels to evaluate the effectiveness of T$^2$HyDHZ, revealing that it is highly insensitive to random haze degrees, as illustrated
in Figure 14.    
In summary, T$^2$HyDHZ exhibits lower standard deviations than the state-of-the-art (SOTA) baseline, indicating better robustness for HyDHZ, as shown in Figure 14.}
%
%
%

%
\subsection{Ablation Study}
\label{sec:ablation}
\begin{table}[t]
\caption{Ablation study for the proposed T$^2$HyDHZ algorithm, aiming to evaluate the efficacy of its constituent components in isolation.}
\renewcommand\arraystretch{1.2}
\vspace{-0.1cm}
\begin{center}
\setlength{\tabcolsep}{1.2mm}
\scalebox{1.13}{
\begin{tabular}{c c c c |c c c c}
\hline
\hline
\makecell[c]{ABS} & SR & \makecell[c]{SSE \\ (Spectral)} & \makecell[c]{SSE \\ (Spatial)} & PSNR & UIQI & SAM & SSIM

\\
\hline
\checkmark  &\checkmark  & &  & 40.391 & 0.983 & 2.741 & 0.994         
\\
\hline
 & \checkmark  & \checkmark & \checkmark & 44.843 & 0.995 & 1.378 & 0.998
 
\\ 
\hline
\checkmark& \checkmark& \checkmark &   & 44.828 &0.995& 1.222 & 0.998

\\

\hline

\checkmark & \checkmark &  & \checkmark& 44.596 & 0.993  & 1.395  & 0.998

\\ 
\hline
\checkmark & \checkmark& \checkmark    & \checkmark& \textbf{{46.102}}  & \textbf{{0.996}}  & \textbf{{1.062}}  & \textbf{{0.999}}  

\\ 
\hline
\hline
\end{tabular}
}
\label{tab:ablation study}
\end{center}
\end{table}
%
%
{As demonstrated in Sections \ref{sec: Qualitative Results} and \ref{sec:quantitativeassessment}, T$^2$HyDHZ has shown SOTA performance in both qualitative and quantitative analyses.
To further analyze the effectiveness of the T$^2$HyDHZ algorithm, we conducted ablation studies as introduced below.
In the following paragraphs, we will first review the significance of designed modules and their respective performance as outlined in Table \ref{tab:ablation study}.  
Afterward, we will examine the effectiveness of the designed $L_\text{sparsity}$ loss as summarized in Table \ref{table:loss}.
Finally, we will demonstrate the benefits of concatenation component within the framework, as illustrated in Table \ref{table:concat}.}
%

%
In Figure \ref{fig:weight}, the ABS block captures the informative bands implied in the original hazy HSI, whose effectiveness can be verified in Table \ref{tab:ablation study}.
Through the second and last rows of Table \ref{tab:ablation study}, we can discover that the efficacy of the ABS block enhances not only spectral recovery with the lowest SAM but also spatial quality with the highest PSNR.
This fact highlights the importance of extracting significant spatial messages {from} essential bands and validates the effectiveness of the ABS block in the proposed framework.
{After} the SR block, the function of the SSE module is to further refine the spectrally super-resolved solution obtained from the ABS and SR blocks.
Therefore, to evaluate the individual performance of the SSE block, we consider the corresponding version of ABS+SR blocks \textit{with/without} the SSE refinement block.
As evident from the first and last rows of Table \ref{tab:ablation study}, the ABS and SR blocks with the SSE refinement yield state-of-the-art performance, demonstrating noticeable advancement in all metrics.
Finally, we evaluate the efficacy of the proposed SSE block, which jointly exploits spatial and spectral information to restore image details.
{Theoretically}, if we just consider the spectral or spatial direction in the SSE module, the data will be only optimized in given dimensions, leading to limited performance.
In this experiment, we evaluate the proposed SSE block  (i.e., $\text{SSE}(\cdot)$) by building the refinement block with only spatial or only spectral transformer layer (i.e., $\text{SpaR}(\cdot)$ or $\text{SpeR}(\cdot)$) instead of an alternate spatial-spectral transformer structure. 
As proven in Table \ref{tab:ablation study}, the encoder that only considers spatial or spectral information performs less than one that considers both, which shows the effectiveness of our designed spatial-spectral transformer-based refinement module.
{To summarize, connecting the spectral and spatial attention encoders through an alternative mechanism achieves improved spectral and spatial image quality, which is demonstrated by the increased PSNR and SAM metrics.}

{As introduced in Section \ref{sec:loss}, we train the T$^2$HyDHZ by further considering the $L_\text{sparsity}$ loss as it aligns with the natural physical property that visible bands with lower transmission rates are more sensitive to haze interference.
Furthermore, to assess the impact of the designed $L_\text{sparsity}$ loss, we have conducted an ablation study \textit{with/without} $L_\text{sparsity}$ loss.
In summary, the results in Table \ref{table:loss} demonstrate that incorporating $L_\text{sparsity}$ loss leads to improved quantitative performance that aligns well with the design logic using natural physical characteristics.
Besides, for evaluating the benefits of the concatenation component introduced in Section \ref{sec:ipt}, we conducted ablation studies to assess its effectiveness.
Specifically, three cases were tested: one is the case without a concatenation component (i.e., the first row of Table \ref{table:concat}), another is the case concatenated with hazy HSI $\bY$ (i.e., the second row of Table \ref{table:concat}), and the other is the case concatenated with weighted MSI $\bY_s$ (i.e., the third row of Table \ref{table:concat}).
The first case evaluates the importance of residual structure, while the second and third assess the performance difference between the modified and original IPT.
Not surprisingly, Table \ref{table:concat} indicates that the inclusion of the commonly used residual structure has resulted in improved performance.
Furthermore, the second case significantly improves the performance in PSNR, UIQI, and SSIM for T$^2$HyDHZ than the third case, demonstrating the efficacy of our model design. 
In general, the performance using $\bY_s$ and $\bY$ both perform rather good performance, showing $\bY_s$ is indeed the representative bands of $\bY$ as a side proof.}

\begin{table}[t]

\centering
\caption{Quantitative evaluation of the proposed loss function with/without $L_\text{sparsity}$ loss.}
\label{table:loss}	
\setlength{\tabcolsep}{7mm}{
\tabcolsep0.05in
\scalebox{0.82}{
 \renewcommand\arraystretch{1.2}
\resizebox{\linewidth}{!}{
\begin{tabular}{ccccc}
\hline
Loss&PSNR&UIQI&SAM&SSIM\\
\hline
${L}_\text{rMRAE}+ {L}_\text{sparsity}$&\bf 46.102&\bf 0.996&\bf 1.062&\bf 0.999\\
${L}_\text{rMRAE}$&45.780&\bf 0.996&\bf 1.062&\bf 0.999\\
\hline
\end{tabular}
}
}
}
\end{table}

\begin{table}[t]

\centering
\caption{Quantitative evaluation of T$^2$HyDHZ with/without concatenation component. The first row corresponds to the case without concatenation; the second and third rows correspond to the case concatenated with hazy HSI $\bY$ and weighted MSI $\bY_s$, respectively.}
\label{table:concat}	
\setlength{\tabcolsep}{7mm}{
\tabcolsep0.05in
\scalebox{0.75}{
 \renewcommand\arraystretch{1.2}
\resizebox{\linewidth}{!}{
\begin{tabular}{ccccc}
\hline
Concatenation&PSNR&UIQI&SAM&SSIM\\
\hline
-&40.266&0.980&1.917&0.992\\
$\bY$&\bf46.102&\bf 0.996&1.062&\bf 0.999\\
$\bY_s$&45.686&0.995&\bf1.033&0.998\\
\hline
\end{tabular}
}
}
}
\end{table}
\section{Conclusion}\label{sec:conclusion}

We elegantly transform the hyperspectral dehazing (HyDHZ) problem into the spectral super-resolution problem{, motivated by} the IPT theory \cite{iptwhsipers} {with interpretable physical meaning}, thereby efficiently solving it with a customized deep transformer network.
As it turns out, the developed T$^2$HyDHZ algorithm is fully automatic and several orders of magnitude faster than existing benchmark methods.
The powerful spatial-spectral transformer is employed for the HyDHZ problem for the first time, allowing us to effectively {select from the haze-corrupted HSI some informative spectral
bands} that are then super-resolved/refined to obtain the target clean hyperspectral data.
Most importantly, without requiring the users to manually set parameters or to mark the hazy areas, the proposed T$^2$HyDHZ algorithm blindly achieves {SOTA} quantitative and qualitative HyDHZ performances.
Therefore, we have developed a highly user-friendly, very fast, and high-performance T$^2$HyDHZ algorithm for facilitating the subsequent hyperspectral remote sensing tasks.

\bibliographystyle{IEEEtran}
\bibliography{refs}

\begin{thebibliography}{10}
\providecommand{\url}[1]{#1}
\csname url@samestyle\endcsname
\providecommand{\newblock}{\relax}
\providecommand{\bibinfo}[2]{#2}
\providecommand{\BIBentrySTDinterwordspacing}{\spaceskip=0pt\relax}
\providecommand{\BIBentryALTinterwordstretchfactor}{4}
\providecommand{\BIBentryALTinterwordspacing}{\spaceskip=\fontdimen2\font plus
\BIBentryALTinterwordstretchfactor\fontdimen3\font minus
  \fontdimen4\font\relax}
\providecommand{\BIBforeignlanguage}[2]{{%
\expandafter\ifx\csname l@#1\endcsname\relax
\typeout{** WARNING: IEEEtran.bst: No hyphenation pattern has been}%
\typeout{** loaded for the language `#1'. Using the pattern for}%
\typeout{** the default language instead.}%
\else
\language=\csname l@#1\endcsname
\fi
#2}}
\providecommand{\BIBdecl}{\relax}
\BIBdecl

\bibitem{BPCSNC2013}
J.~M. Bioucas-Dias, A.~Plaza, G.~Camps-Valls, P.~Scheunders, N.~Nasrabadi, and
  J.~Chanussot, ``Hyperspectral remote sensing data analysis and future
  challenges,'' \emph{IEEE Geoscience and Remote Sensing Magazine}, vol.~1,
  no.~2, pp. 6--36, Jun. 2013.

\bibitem{reviewaccess14}
M.~J. Khan, H.~S. Khan, A.~Yousaf, K.~Khurshid, and A.~Abbas, ``Modern trends
  in hyperspectral image analysis: A review,'' \emph{IEEE Access}, vol.~6, pp.
  14\,118--14\,129, Mar. 2018.

\bibitem{lin2021all}
C.-H. Lin and T.-H. Lin, ``All-addition hyperspectral compressed sensing for
  metasurface-driven miniaturized satellite,'' \emph{IEEE Transactions on
  Geoscience and Remote Sensing}, vol.~60, pp. 1--15, Mar. 2021.

\bibitem{Hongtgrs2021}
D.~Hong, L.~Gao, J.~Yao, B.~Zhang, A.~Plaza, and J.~Chanussot, ``Graph
  convolutional networks for hyperspectral image classification,'' \emph{IEEE
  Transactions on Geoscience and Remote Sensing}, vol.~59, no.~7, pp.
  5966--5978, Jul. 2021.

\bibitem{Houtgrs2022}
Z.~Hou, W.~Li, L.~Li, R.~Tao, and Q.~Du, ``Hyperspectral change detection based
  on multiple morphological profiles,'' \emph{IEEE Transactions on Geoscience
  and Remote Sensing}, vol.~60, pp. 1--12, Jul. 2021.

\bibitem{codehcd}
T.-H. Lin and C.-H. Lin, ``Hyperspectral change detection using semi-supervised
  graph neural network and convex deep learning,'' \emph{IEEE Transactions on
  Geoscience and Remote Sensing}, vol.~61, pp. 1--18, Jun. 2023.

\bibitem{underwood2006mapping}
E.~Underwood, M.~Mulitsch, J.~Greenberg, M.~Whiting, S.~Ustin, and S.~Kefauver,
  ``Mapping invasive aquatic vegetation in the {S}acramento-{S}an {J}oaquin
  {D}elta using hyperspectral imagery,'' \emph{Environmental Monitoring and
  Assessment}, vol. 121, no.~1, pp. 47--64, Jun. 2006.

\bibitem{codemm}
C.-H. Lin, M.-C. Chu, and P.-W. Tang, ``{CODE-MM}: Convex deep mangrove mapping
  algorithm based on optical satellite images,'' \emph{IEEE Transactions on
  Geoscience and Remote Sensing}, vol.~61, pp. 1--19, Sep. 2023.

\bibitem{CVXbookCLL2016}
C.-Y. Chi, W.-C. Li, and C.-H. Lin, \emph{Convex Optimization for Signal
  Processing and Communications: {F}rom Fundamentals to Applications}.\hskip
  1em plus 0.5em minus 0.4em\relax CRC Press, Boca Raton, FL, 2017.

\bibitem{narasimhan2002vision}
S.~G. Narasimhan and S.~K. Nayar, ``Vision and the atmosphere,''
  \emph{International Journal of Computer Vision}, vol.~48, no.~3, pp.
  233--254, Jul. 2002.

\bibitem{GuoTGRS2021}
J.~Guo, J.~Yang, H.~Yue, C.~Hou, and K.~Li, ``Landsat-8 {OLI} multispectral
  image dehazing based on optimized atmospheric scattering model,'' \emph{IEEE
  Transactions on Geoscience and Remote Sensing}, vol.~59, no.~12, pp.
  10\,255--10\,265, Dec. 2021.

\bibitem{XudongTGRS2021}
X.~Kang, Z.~Fei, P.~Duan, and S.~Li, ``Fog model-based hyperspectral image
  defogging,'' \emph{IEEE Transactions on Geoscience and Remote Sensing},
  vol.~60, pp. 1--12, Aug. 2021.

\bibitem{DOS2022}
Z.~Li, P.~Duan, S.~Hu, M.~Li, and X.~Kang, ``Fast hyperspectral image dehazing
  with dark-object subtraction model,'' \emph{IEEE Geoscience and Remote
  Sensing Letters}, vol.~19, pp. 1--5, Oct. 2022.

\bibitem{asmodel2000}
S.~Narasimhan and S.~Nayar, ``Chromatic framework for vision in bad weather,''
  in \emph{Proc. IEEE/CVF Conference on Computer Vision and Pattern
  Recognition}, South Carolina, USA, Jun. 13-15, 2000, pp. 598--605.

\bibitem{Caidehazenettip2016}
B.~Cai, X.~Xu, K.~Jia, C.~Qing, and D.~Tao, ``{DehazeNet}: An end-to-end system
  for single image haze removal,'' \emph{IEEE Transactions on Image
  Processing}, vol.~25, no.~11, pp. 5187--5198, Nov. 2016.

\bibitem{LinTNNLS2023}
C.-H. Lin, Y.~Liu, C.-Y. Chi, C.-C. Hsu, H.~Ren, and T.~Q.~S. Quek,
  ``Hyperspectral tensor completion using low-rank modeling and convex
  functional analysis,'' \emph{IEEE Transactions on Neural Networks and
  Learning Systems}, pp. 1--15, Feb. 2023.

\bibitem{hyperqueen}
C.-H. Lin and Y.-Y. Chen, ``Hyper{QUEEN}: Hyperspectral quantum deep network
  for image restoration,'' \emph{IEEE Transactions on Geoscience and Remote
  Sensing}, vol.~61, pp. 1--20, May 2023.

\bibitem{hetpami2010}
K.~He, J.~Sun, and X.~Tang, ``Single image haze removal using dark channel
  prior,'' \emph{IEEE Transactions on Pattern Analysis and Machine
  Intelligence}, vol.~33, no.~12, pp. 2341--2353, Dec. 2011.

\bibitem{TanCVPR2008}
R.~T. Tan, ``Visibility in bad weather from a single image,'' in \emph{Proc.
  IEEE/CVF Computer Vision and Pattern Recognition Conference}, Alaska, USA,
  Jun. 23-28, 2008, pp. 1--8.

\bibitem{bp}
R.~Hecht-Nielsen, ``Theory of the backpropagation neural network,'' in
  \emph{Proc. International 1989 Joint Conference on Neural Networks},
  Washington, USA, Jun. 18-22, 1989, pp. 593--605.

\bibitem{guo2020rsdehazenet}
J.~Guo, J.~Yang, H.~Yue, H.~Tan, C.~Hou, and K.~Li, ``{RSDehazeNet}: Dehazing
  network with channel refinement for multispectral remote sensing images,''
  \emph{IEEE Transactions on Geoscience and Remote Sensing}, vol.~59, no.~3,
  pp. 2535--2549, Mar. 2021.

\bibitem{ma2022spectral}
X.~Ma, Q.~Wang, and X.~Tong, ``A spectral grouping-based deep learning model
  for haze removal of hyperspectral images,'' \emph{ISPRS Journal of
  Photogrammetry and Remote Sensing}, vol. 188, pp. 177--189, Jun. 2022.

\bibitem{lin2016fast}
C.-H. Lin, C.-Y. Chi, Y.-H. Wang, and T.-H. Chan, ``A fast hyperplane-based
  minimum-volume enclosing simplex algorithm for blind hyperspectral
  unmixing,'' \emph{IEEE Transactions on Signal Processing}, vol.~64, no.~8,
  pp. 1946--1961, Apr. 2016.

\bibitem{QinJSTARS2018}
M.~Qin, F.~Xie, W.~Li, Z.~Shi, and H.~Zhang, ``Dehazing for multispectral
  remote sensing images based on a convolutional neural network with the
  residual architecture,'' \emph{IEEE Journal of Selected Topics in Applied
  Earth Observations and Remote Sensing}, vol.~11, no.~5, pp. 1645--1655, May
  2018.

\bibitem{song2022rethinking}
Y.~Song, Y.~Zhou, H.~Qian, and X.~Du, ``Rethinking performance gains in image
  dehazing networks,'' \emph{arXiv:2209.11448}, 2022.

\bibitem{MTCRN2021jstar}
Y.~Zi, F.~Xie, N.~Zhang, Z.~Jiang, W.~Zhu, and H.~Zhang, ``Thin cloud removal
  for multispectral remote sensing images using convolutional neural networks
  combined with an imaging model,'' \emph{IEEE Journal of Selected Topics in
  Applied Earth Observations and Remote Sensing}, vol.~14, pp. 3811--3823, Mar,
  2021.

\bibitem{gan2016dehazing}
Y.~Gan, B.~Hu, D.~Wen, and S.~Wang, ``Dehazing method for hyperspectral remote
  sensing imagery with hyperspectral linear unmixing,'' in \emph{Proc.
  International Symposium on Optoelectronic Technology and Application},
  Beijing, China, May 9-11, 2016, pp. 296--301.

\bibitem{HISUN}
C.-H. Lin and J.~M. Bioucas-Dias, ``Nonnegative blind source separation for
  ill-conditioned mixtures via {J}ohn ellipsoid,'' \emph{IEEE Transactions on
  Neural Networks and Learning Systems}, vol.~32, no.~5, pp. 2209--2223, May
  2021.

\bibitem{lin2015identifiability}
C.-H. Lin, W.-K. Ma, W.-C. Li, C.-Y. Chi, and A.~Ambikapathi, ``Identifiability
  of the simplex volume minimization criterion for blind hyperspectral
  unmixing: {T}he no-pure-pixel case,'' \emph{IEEE Transactions on Geoscience
  and Remote Sensing}, vol.~53, no.~10, pp. 5530--5546, Oct. 2015.

\bibitem{FUNtgrs15}
R.~Guerra, L.~Santos, S.~López, and R.~Sarmiento, ``A new fast algorithm for
  linearly unmixing hyperspectral images,'' \emph{IEEE Transactions on
  Geoscience and Remote Sensing}, vol.~53, no.~12, pp. 6752--6765, Dec. 2015.

\bibitem{Tang2022dehazing}
P.-W. Tang and C.-H. Lin, ``Hyperspectral dehazing using {ADMM-Adam} theory,''
  in \emph{Proc. IEEE Workshop on Hyperspectral Imaging and Signal Processing:
  Evolution in Remote Sensing}, Rome, Italy, Sep. 13-16, 2022, pp. 1--5.

\bibitem{LinTGRS2021}
C.-H. Lin, Y.-C. Lin, and P.-W. Tang, ``{ADMM-ADAM}: A new inverse imaging
  framework blending the advantages of convex optimization and deep learning,''
  \emph{IEEE Transactions on Geoscience and Remote Sensing}, vol.~60, pp.
  1--16, Sep. 2021.

\bibitem{MA2022113012}
X.~Ma, Q.~Wang, X.~Tong, and P.~M. Atkinson, ``A deep learning model for
  incorporating temporal information in haze removal,'' \emph{Remote Sensing of
  Environment}, vol. 274, p. 113012, Jun. 2022.

\bibitem{iptwhsipers}
C.-H. Lin and P.-W. Tang, ``Inverse problem transform: Solving hyperspectral
  inpainting via deterministic compressed sensing,'' in \emph{Proc. IEEE
  Workshop on Hyperspectral Imaging and Signal Processing: Evolution in Remote
  Sensing}, Amsterdam, Netherlands, Mar. 24-26, 2021, pp. 1--5.

\bibitem{Linnature2023}
C.-H. Lin, S.-H. Huang, T.-H. Lin, and P.~C. Wu, ``Metasurface-empowered
  snapshot hyperspectral imaging with convex/deep ({CODE}) small-data learning
  theory,'' \emph{Nature Communications}, vol.~14, Nov. 2023.

\bibitem{Hetnnls2022}
J.~He, J.~Li, Q.~Yuan, H.~Shen, and L.~Zhang, ``Spectral response
  function-guided deep optimization-driven network for spectral
  super-resolution,'' \emph{IEEE Transactions on Neural Networks and Learning
  Systems}, vol.~33, no.~9, pp. 4213--4227, Sep. 2022.

\bibitem{Hangtip2021}
R.~Hang, Q.~Liu, and Z.~Li, ``Spectral super-resolution network guided by
  intrinsic properties of hyperspectral imagery,'' \emph{IEEE Transactions on
  Image Processing}, vol.~30, pp. 7256--7265, Aug. 2021.

\bibitem{Yitgrs2019}
C.~Yi, Y.-Q. Zhao, and J.~C.-W. Chan, ``Spectral super-resolution for
  multispectral image based on spectral improvement strategy and spatial
  preservation strategy,'' \emph{IEEE Transactions on Geoscience and Remote
  Sensing}, vol.~57, no.~11, pp. 9010--9024, Nov. 2019.

\bibitem{NIPS2015_8fb21ee7}
S.~Sukhbaatar, A.~Szlam, J.~Weston, and R.~Fergus, ``End-to-end memory
  networks,'' in \emph{Proc. International Conference on Neural Information
  Processing Systems}, Quebec, Canada, Dec. 7-12, 2015, pp. 1--9.

\bibitem{nair2010rectified}
V.~Nair and G.~E. Hinton, ``Rectified linear units improve restricted
  {B}oltzmann machines,'' in \emph{Proc. International Conference on Machine
  Learning}, Haifa, Israel, Jun. 21-24, 2010, pp. 807--814.

\bibitem{Hendrycks2016GaussianEL}
\BIBentryALTinterwordspacing
D.~Hendrycks and K.~Gimpel, ``Gaussian error linear units ({GELUs}),''
  \emph{Computing Research Repository}, vol. abs/1606.08415, 2016. [Online].
  Available: \url{http://arxiv.org/abs/1606.08415}
\BIBentrySTDinterwordspacing

\bibitem{attentionisallyoueed}
A.~Vaswani, N.~Shazeer, N.~Parmar, J.~Uszkoreit, L.~Jones, A.~N. Gomez,
  L.~Kaiser, and I.~Polosukhin, ``Attention is all you need,'' in \emph{Proc.
  International Conference on Neural Information Processing Systems},
  California, USA, Dec. 4-9, 2017, p. 6000–6010.

\bibitem{wang2020cnn}
Z.~J. Wang, R.~Turko, O.~Shaikh, H.~Park, N.~Das, F.~Hohman, M.~Kahng, and
  D.~H.~P. Chau, ``{CNN} explainer: {L}earning convolutional neural networks
  with interactive visualization,'' \emph{IEEE Transactions on Visualization
  and Computer Graphics}, vol.~27, no.~2, pp. 1396--1406, Feb. 2021.

\bibitem{DCSN}
C.-C. Hsu, C.-H. Lin, C.-H. Kao, and Y.-C. Lin, ``{DCSN}: Deep compressed
  sensing network for efficient hyperspectral data transmission of miniaturized
  satellite,'' \emph{IEEE Transactions on Geoscience and Remote Sensing},
  vol.~59, no.~9, pp. 7773--7789, Sep. 2020.

\bibitem{mraetgrs}
J.~Li, C.~Wu, R.~Song, W.~Xie, C.~Ge, B.~Li, and Y.~Li, ``Hybrid 2-{D}–3-{D}
  deep residual attentional network with structure tensor constraints for
  spectral super-resolution of {RGB} images,'' \emph{IEEE Transactions on
  Geoscience and Remote Sensing}, vol.~59, no.~3, pp. 2321--2335, Mar. 2021.

\bibitem{shi2018deep}
Z.~Shi, C.~Chen, Z.~Xiong, D.~Liu, Z.-J. Zha, and F.~Wu, ``Deep residual
  attention network for spectral image super-resolution,'' in \emph{Proc. of
  the European Conference on Computer Vision Workshops}, Munich, Germany, Sep.
  8-14, 2018, pp. 214--229.

\bibitem{kingma2014adam}
D.~P. Kingma and J.~Ba, ``{ADAM}: A method for stochastic optimization,'' in
  \emph{Proc. IEEE International Conference on Learning Representations}, San
  Diego, USA, May 7-9, 2015, pp. 1--15.

\bibitem{portal}
``\textit{AVIRIS {D}ata {P}ortal},'' {A}ccessed: Nov. 2, 2019. [Online].
  Available: \url{https://aviris.jpl.nasa.gov/dataportal/}.

\bibitem{mccartney1976optics}
E.~J. McCartney, \emph{Optics of the Atmosphere: {S}cattering by {M}olecules
  and {P}articles}.\hskip 1em plus 0.5em minus 0.4em\relax John Wiley and Sons,
  Inc., Hoboken, New Jersey, 1976.

\bibitem{chavez1988improved}
P.~S. Chavez~Jr., ``An improved dark-object subtraction technique for
  atmospheric scattering correction of multispectral data,'' \emph{Remote
  Sensing of Environment}, vol.~24, no.~3, pp. 459--479, Apr. 1988.

\bibitem{psnr}
A.~Tanchenko, ``Visual-{PSNR} measure of image quality,'' \emph{Journal of
  Visual Communication and Image Representation}, vol.~25, no.~5, pp. 874--878,
  Jul. 2014.

\bibitem{UIQI2002}
Z.~Wang and A.~C. Bovik, ``A universal image quality index,'' \emph{IEEE Signal
  Processing Letters}, vol.~9, no.~3, pp. 81--84, Mar. 2002.

\bibitem{SAM92}
R.~H. Yuhas, A.~F. Goetz, and J.~W. Boardman, ``Discrimination among semi-arid
  landscape endmembers using the spectral angle mapper {(SAM)} algorithm,'' in
  \emph{Summaries of the Third Annual JPL Airborne Geoscience Workshop},
  California, USA, Jun. 1-5 1992, pp. 147--149.

\bibitem{lin2017COCNMF}
C.-H. Lin, F.~Ma, C.-Y. Chi, and C.-H. Hsieh, ``A convex optimization-based
  coupled nonnegative matrix factorization algorithm for hyperspectral and
  multispectral data fusion,'' \emph{IEEE Transactions on Geoscience and Remote
  Sensing}, vol.~56, no.~3, pp. 1652--1667, Mar. 2018.

\bibitem{ssim}
Z.~Wang, A.~Bovik, H.~Sheikh, and E.~Simoncelli, ``Image quality assessment:
  {F}rom error visibility to structural similarity,'' \emph{IEEE Transactions
  on Image Processing}, vol.~13, no.~4, pp. 600--612, Apr. 2004.

\end{thebibliography}
\begin{IEEEbiography}[{\resizebox{1in}{!}{\includegraphics[width=1in,height=1.25in,clip,keepaspectratio]{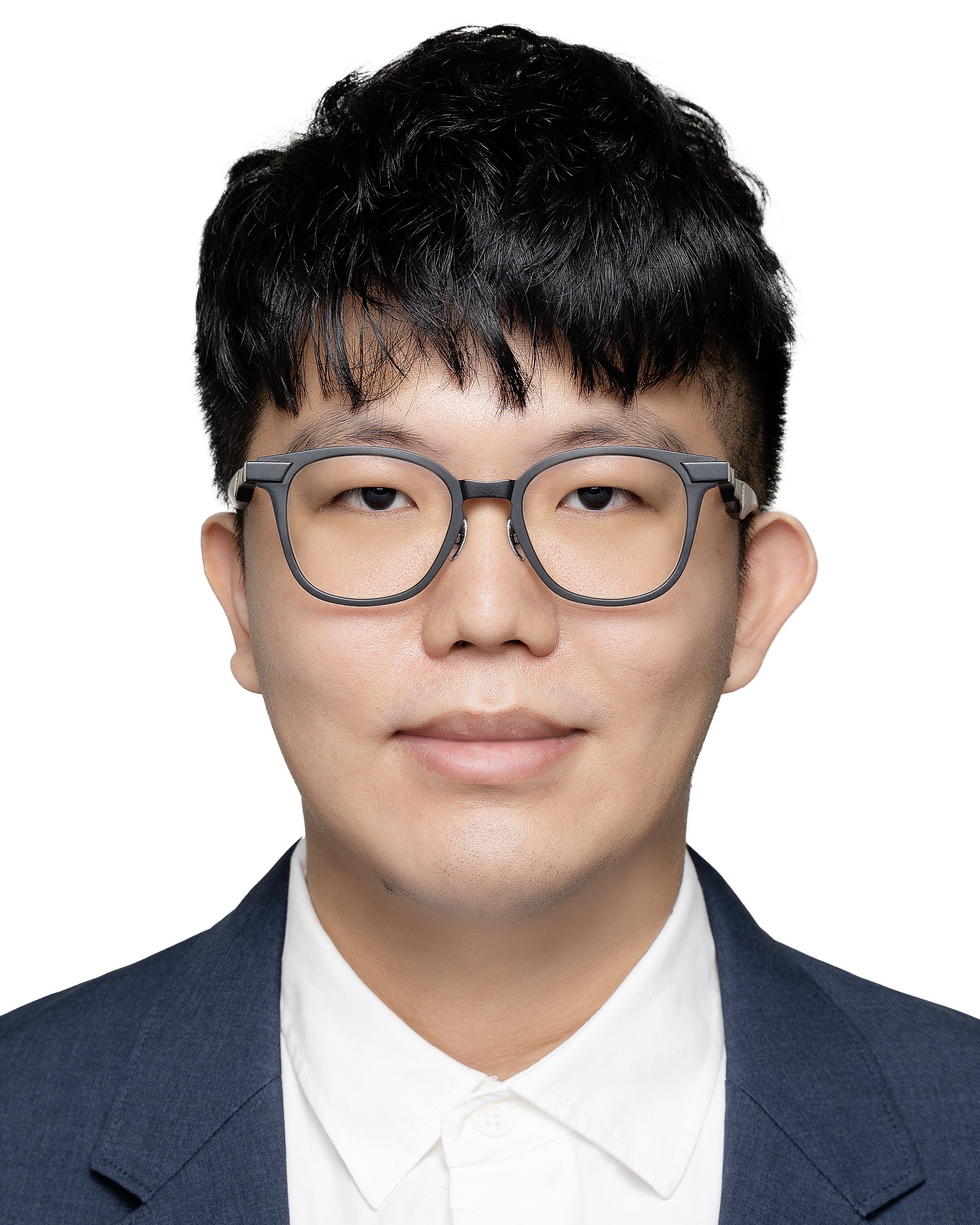}}}]
		{\bf Po-Wei Tang}
    (S'20)
received his B.S. degree from the Department of Electronic Engineering, National Changhua University of Education, Taiwan, in 2018.
He is currently a Ph.D. student with the Intelligent Hyperspectral Computing Laboratory, Institute of Computer and Communication Engineering, National Cheng Kung University (NCKU), Taiwan. 
His research interests include deep learning, convex optimization, tensor completion, and hyperspectral imaging.
He has been selected as a recipient of the Graduate Students Study Abroad Program from the National Science and Technology Council (NSTC).
He received a highly competitive scholarship from NCKU, as well as the Pan Wen Yuan Award from the Industrial Technology Research Institute (ITRI) of Taiwan.
	\end{IEEEbiography}
%
\begin{IEEEbiography}[{\resizebox{0.9in}{!}{\includegraphics[width=1in,height=1.25in,clip,keepaspectratio]{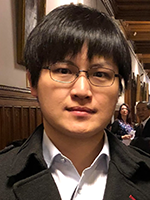}}}]
	{\bf Chia-Hsiang Lin}
	(S'10-M'18)
received the B.S. degree in electrical engineering and the Ph.D. degree in communications engineering from National Tsing Hua University (NTHU), Taiwan, in 2010 and 2016, respectively.
From 2015 to 2016, he was a Visiting Student of Virginia Tech,
Arlington, VA, USA.

He is currently an Associate Professor with the Department of Electrical Engineering, and also with 
the Miin Wu School of Computing,
National Cheng Kung University (NCKU), Taiwan.
Before joining NCKU, he held research positions with The Chinese University of Hong Kong, HK (2014 and 2017), 
NTHU (2016-2017), 
and the University of Lisbon (ULisboa), Lisbon, Portugal (2017-2018).
He was an Assistant Professor with the Center for Space and Remote Sensing Research, National Central University, Taiwan, in 2018, and a Visiting Professor with ULisboa, in 2019.
His research interests include network science, 
quantum computing,
convex geometry and optimization, blind signal processing, and imaging science.

Dr. Lin received the Emerging Young Scholar Award from National Science and Technology Council (NSTC), in 2023,
the Future Technology Award from NSTC, in 2022,
the Outstanding Youth Electrical Engineer Award from The Chinese Institute of Electrical Engineering (CIEE), in 2022,
the Best Young Professional Member Award from IEEE Tainan Section, in 2021,
the Prize Paper Award from IEEE Geoscience and Remote Sensing Society (GRS-S), in 2020, 
the Top Performance Award from Social Media Prediction Challenge at ACM Multimedia, in 2020,
and The 3rd Place from AIM Real World Super-Resolution Challenge at IEEE International Conference on Computer Vision (ICCV), in 2019. 
He received the Ministry of Science and Technology (MOST) Young Scholar Fellowship, together with the EINSTEIN Grant Award, from 2018 to 2023.
In 2016, he was a recipient of the Outstanding Doctoral Dissertation Award from the Chinese Image Processing and Pattern Recognition Society and the Best Doctoral Dissertation Award from the IEEE GRS-S.
    \end{IEEEbiography}

    \begin{IEEEbiography}[{\resizebox{1in}{!}{\includegraphics[width=1in,height=1.25in,clip,keepaspectratio]{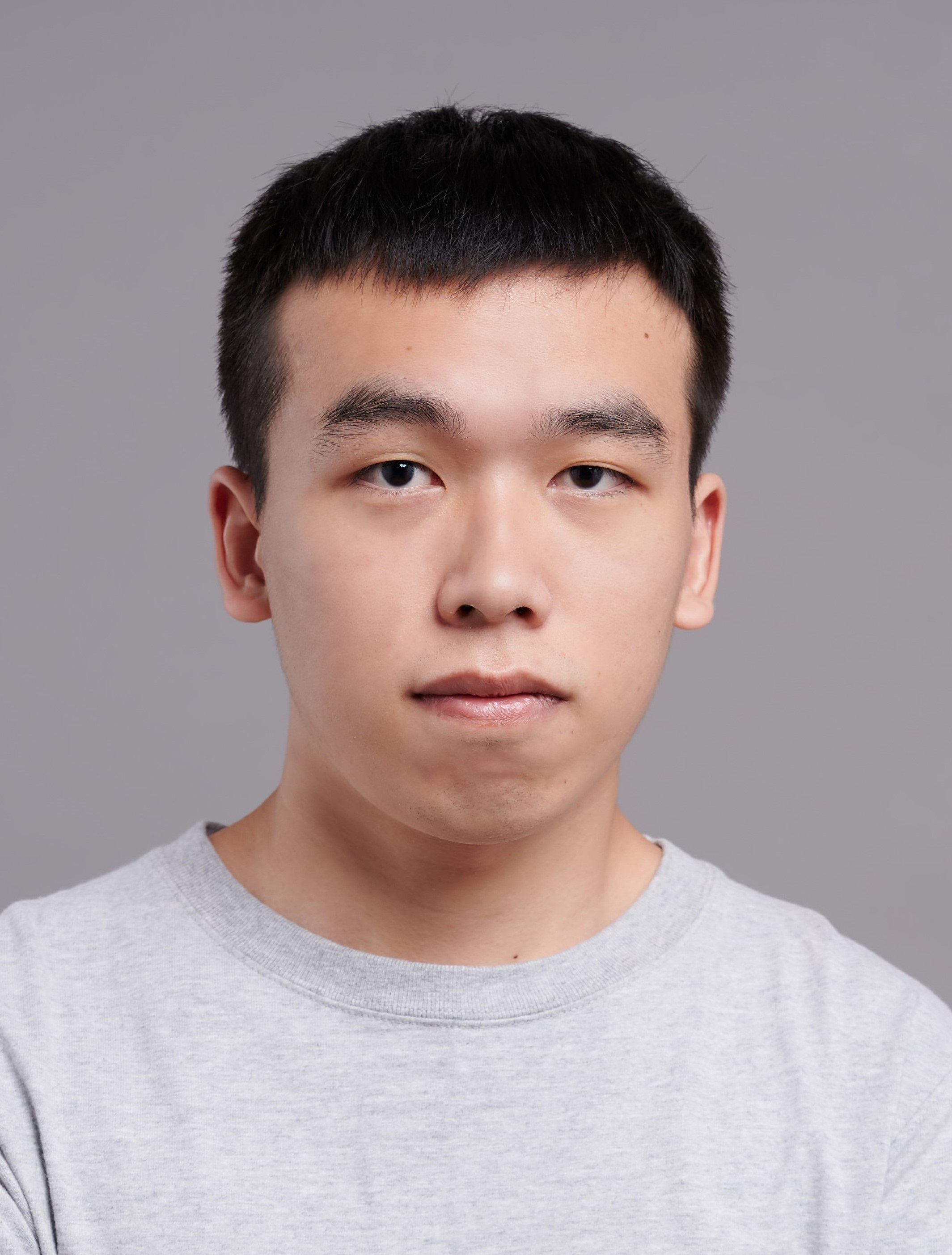}}}]{\bf Yangrui Liu}
(S’20)
received his B.S. degree from the Department of Communications Engineering, Feng Chia University, Taiwan, in 2019.
He is currently a Ph.D. student with Intelligent Hyperspectral Computing Laboratory, Institute of Computer and Communication Engineering, National Cheng Kung University, Taiwan. 
His research interests include convex optimization, deep learning, and hyperspectral anomaly detection.
\end{IEEEbiography}

\end{document}